\documentclass[letterpaper]{article} 
\usepackage[draft]{aaai2026}  
\usepackage{times}  
\usepackage{helvet}  
\usepackage{courier}  
\usepackage[hyphens]{url}  
\usepackage{graphicx} 
\usepackage{natbib}  
\usepackage{caption} 
\usepackage{algorithm}
\usepackage{algorithmic}

\usepackage{newfloat}
\usepackage{listings}
\DeclareCaptionStyle{ruled}{labelfont=normalfont,labelsep=colon,strut=off} 
\floatstyle{ruled}
\newfloat{listing}{tb}{lst}{}
\floatname{listing}{Listing}
\usepackage{amsthm}
\usepackage{amsmath}
\usepackage{amsfonts}       
\usepackage{subfigure}
\usepackage{array}
\usepackage{booktabs}       
\usepackage{makecell}

\usepackage{enumitem}

\theoremstyle{plain}
\newtheorem{theorem}{Theorem}[section]
\newtheorem{corollary}{Corollary}[section]
\newtheorem{definition}{Definition}[section]

\renewcommand{\thesubfigure}{(\alph{subfigure})}
\makeatletter \renewcommand{\@thesubfigure}{\thesubfigure \space}
\renewcommand{\p@subfigure}{} \makeatother
\newcommand*\samethanks[1][\value{footnote}]{\footnotemark[#1]}

\title{Towards Revealing the Effectiveness of Small-Scale Fine-Tuning in R1-Style Reinforcement Learning}

\author {
    Yutong Chen\textsuperscript{\rm 1},
    Jiandong Gao\textsuperscript{\rm 1}\thanks{Corresponding author.},
    Ji Wu\textsuperscript{\rm 1, \rm 2}\samethanks
}
\affiliations {
    \textsuperscript{\rm 1}Tsinghua University\\
    \textsuperscript{\rm 2} Beijing National Research Center for Information Science and Technology\\
    chen-yt23@mails.tsinghua.edu.cn, jdgao@tsinghua.edu.cn, wuji\_ee@tsinghua.edu.cn
}

\begin{document}

\maketitle

\begin{abstract}
R1-style Reinforcement Learning (RL) significantly enhances Large Language Models' reasoning capabilities, yet the mechanism behind rule-based RL remains unclear. We found that small-scale SFT has substantial influence on RL but shows poor efficiency. To explain our observations, we propose an analytical framework and compare the efficiency of SFT and RL by measuring \textbf{sample effect}. Our hypothetical analysis shows the potential to improve SFT efficiency. Guided by our analysis, we propose \textbf{Re-distillation}, a technique that aims to boost the effectiveness of small-scale distillation by sampling from the RL-trained policy. 
Re-distillation shows consistent surprising efficiency on three datasets and both Qwen\&Llama models: Re-distilled models matched RL performance with far fewer samples and less computation. As a result, on K\&K dataset, our re-distilled Qwen-2.5-1.5B model surpasses DeepSeek-V3-0324 with only 1K SFT samples. We demonstrate that re-distillation can be used to efficiently balance multiple goals in RL. Our work explains several interesting phenomena in R1-style RL, shedding light on the mechanisms behind its empirical success. Code is available at: \url{https://github.com/on1262/deep-reasoning}
\end{abstract}

\section{Introduction} \label{sec:intro}

Recent advances in Reinforcement Learning for Large Language Models (LLMs) demonstrate remarkable improvements in reasoning-intensive tasks \citep{guo_deepseek-r1_2025,kimi_team_kimi_2025,qwen_team_qwq-32b_2025,openai_openai_2024,openai_competitive_2025}. The R1-style RL, we use this term as outcome-based reinforcement learning without step-wise verification, typically involves two phases: a cold-start Supervised Fine-Tuning(SFT), where models learn instruction following and long Chain-of-Thought \citep{wei_chain--thought_2022} through curated reasoning traces, followed by RL optimization. Empirical evidence suggests improving SFT quality enhances RL performance. For instance, DeepSeek-R1 employed carefully filtered reasoning traces from DeepSeek-R1-Zero \citep{guo_deepseek-r1_2025}. However, the sample selection principle in this SFT stage remains blur. In this paper, we focus on \textbf{small-scale SFT} (\(\leq\)1K samples). \textbf{This limitation ensures performance gains mainly from eliciting reasoning capability rather than simply knows more}, as we assume that knowledge in such small amount of data is negligible compared with enormous pretraining corpus. 

The relationship between SFT and RL seems like a paradox: while SFT provides richer supervision, RL remains essential for further improvement \citep{chu_sft_2025}. Recent works found strong memorization effect during LLM pretraining, recalling rare information from minimal exposure \citep{hartmann_sok_2023,carlini_quantifying_2022}. 
On the contrary, rule-based RL is under severe information constraints. A binary reward function with solution correctness probability \(p\) conveys merely \(H = -plogp - (1-p)log(1-p) \leq 1\) bit on average.

\begin{figure*}[h]
  \centering
  \includegraphics[width=0.8\textwidth]{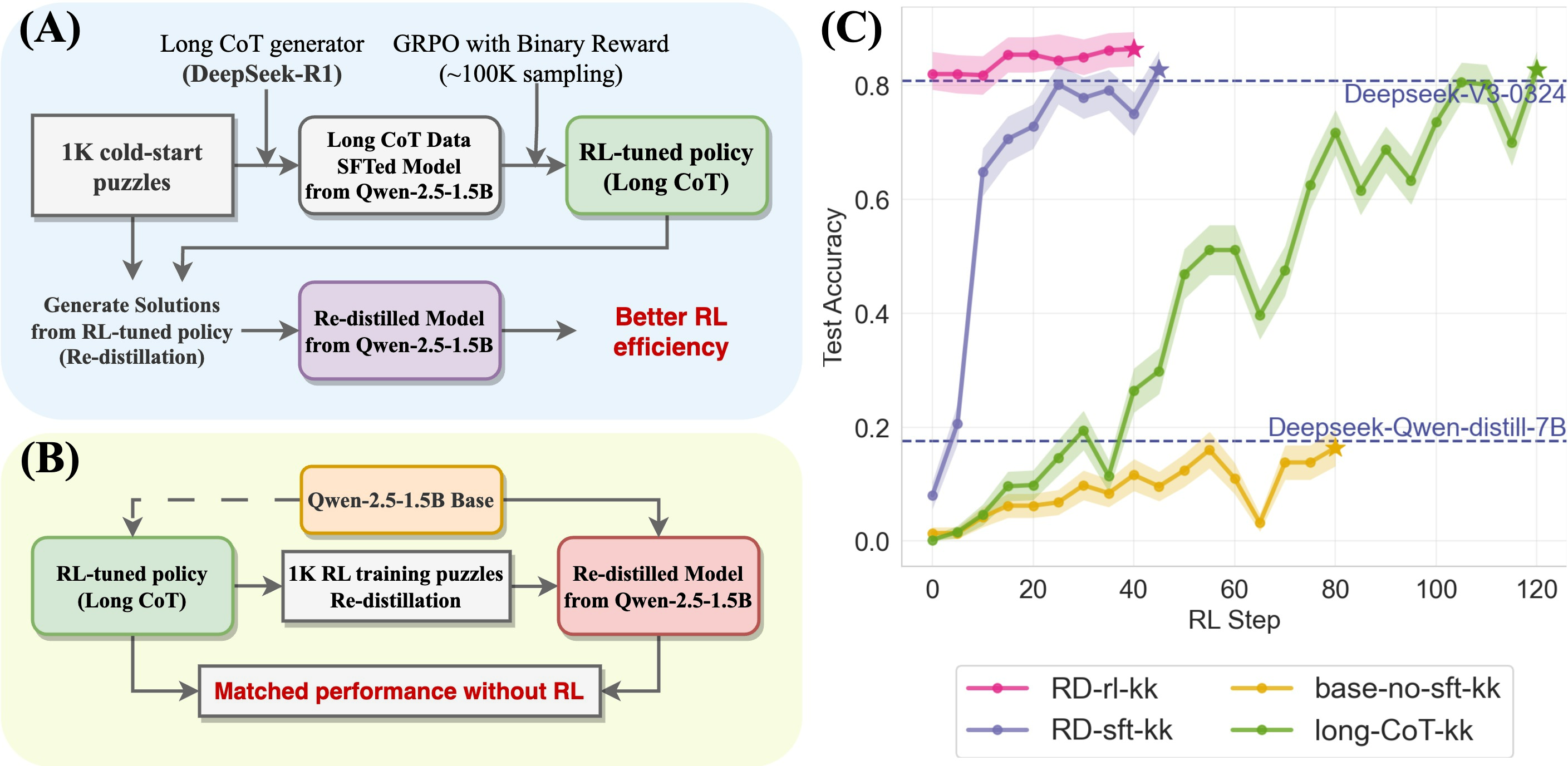}
  \caption{Main experiments overview. \textbf{Left (A)}: We investigate which method produces better samples for small-scale SFT before RL. Pipeline(A) shows samples distilled from RL-trained policy lead to the fastest convergence in RL stage. Common approaches, such as distilling from stronger model or using synthetic solutions, failed to maximize RL efficiency as depicted in (C). \textbf{Left (B)}: It is possible to reproduce RL-trained policy's performance by purely SFT. We randomly pick 1K puzzles from RL training set and generate solutions by the RL-trained policy, then re-distill base model. The re-distilled model matched RL policy immediately after fine-tuning. \textbf{Right (C)}: The RL test accuracy on K\&K dataset. Two re-distilled models(\texttt{RD-sft-kk} and \texttt{RD-rl-kk}) have higher performance with fewer steps. See Section \ref{sec:re-distill} for more experiments and details. Shaded area: 95\% Confidence Interval.}
  \label{fig:teaser}
\end{figure*}

Our research starts from examining SFT efficiency through experiments on K\&K \citep{xie_memorization_2024} and MATH \citep{hendrycks_measuring_2021} dataset. The overview of main experiments is depicted in Figure \ref{fig:teaser}. Firstly, we construct SFT datasets by different methods to investigate if small-scale fine-tuning has substantial influence. Fine-tuning Qwen-2.5-1.5B Base model with \(\approx\) 1K samples reveals that small-scale SFT induces non-negligible changes in RL stage. However, the results can not be simply interpreted by simple data collection principles, because model behaviors are inconsistent on different datasets.

We explain our observations by establishing an analytical framework. Under the hypothesis that non-linear training dynamic of RL can be analyzed by linearized model, the effectiveness of both RL and SFT can be decomposed to each sample's contribution. We name this contribution as \textbf{sample effect}. It is proved that sample effect has great influence on SFT effectiveness. Specifically, the optimal policy for distillation will output samples with high effect\footnote{In this paper, both distillation and re-distillation use hard labels (token IDs). We do not fit token distribution by soft label.}. We also proved that RL improves output sample effect and test performance simultaneously, which suggests the potential of learning from RL-trained policy.

Inspired by hypothetical analysis, we propose the \textbf{Re-distillation} technique to unleash the potential of small-scale SFT. This method samples from converged policies to generate new SFT data. Remarkably, base models trained on re-distilled data match converged RL performance using only SFT, demonstrating superior data efficiency. We verify the effectiveness of re-distillation on three datasets and two models, showing that complex RL process can be `compressed' into SFT process. We also demonstrate an application that merges two RL-trained policies by re-distillation in order to balance different goals.

While distillation is a conventional approach, re-distillation is a special case with both outstanding results and extra requirements. Leveraging this effect requires understanding its deep connection with R1-style RL, which is not fully investigated in prior distillation works \citep{wang2023self,guminillm,pmlr-v202-liang23j,tunstall2023zephyr}.

Current data collection principle for SFT or RL often emphasize data quality from human perspective, such as deep thinking behavior or problem diversity \citep{ye_limo_2025,chu_sft_2025,gandhi_cognitive_2025}. However, these principles have less significant effects or inconsistent results. We offer both theoretical and empirical study to understand those counter-intuitive results, revealing the deep connection between model behavior and data.

We list the main contributions and findings as follows: 

\begin{itemize}[leftmargin=15pt,itemindent=2pt]
\item We demonstrate that SFT can be as generalizable as RL. Fewer than \(1\)K SFT samples reproduces RL's performance. We propose Re-distillation as a \emph{post-hoc} method for creating high-efficiency SFT datasets (Section \ref{sec:re-distill}). Our proposed method has higher efficiency than simply scale up SFT dataset (Subsection \ref{subsec:more-sft})

\item Our hypothetical analysis reveals that the poor sample efficiency is not an intrinsic property of SFT. SFT training dynamic is influenced by \emph{sample effects} (Subsection \ref{subsec:theory-opt-target}). We provide empirical study to verify our hypothesis (Section \ref{sec:exp-verify}).

\item We explain why re-distillation is so effective. Specifically, we proved a lower bound of output sample effect on RL-trained policy which relate to the reward growth rate (Subsection \ref{subsec:theory-rl-effect}). 

\item We explain why SFT influence RL exploration in the long term. Our experiment shows RL's inherent difficulty in modifying initial token distributions compared to SFT {Subsection \ref{subsec:exploration}}.

\item We provide an application which uses re-distillation to merge two RL-trained policies by only SFT. Editing SFT dataset enables efficient balancing multiple RL goals and enhancing model performance for specific downstream tasks (Subsection \ref{subsec:merge}).

\end{itemize}

\section{Related Works}

\textbf{R1-style Reinforcement Learning}: Recent studies demonstrate small-scale SFT's considerable influence on subsequent RL performance. \citet{gandhi_cognitive_2025} shows Qwen and Llama benefit from \(1\)K SFT samples in reasoning tasks. \citet{yeo_demystifying_2025} finds high-quality SFT data is crucial for releasing RL potential. \citet{zeng_simplerl-zoo_2025} reports degraded performance with short CoT data than RL on base models. \citet{zhao_echo_2025} observes RL tends to fall into single pretrain mode. Other common observations include unstable response length growth and the pre-existence of reasoning patterns in base models \citep{zeng_simplerl-zoo_2025,xie_logic-rl_2025,yeo_demystifying_2025}. Recent studies also highlight some important attributes of R1-style RL: \citep{wang_reinforcement_2025} demonstrated that RL can be effective with only one questions. Note that the definition of one-shot RL allows model generate many responses from a single question. \citet{shao_spurious_2025} found that even weak or spurious rewards can also elicit reasoning ability. 

\textbf{Direct SFT Approaches}: Emerging research achieves remarkable results through small-scale SFT. By leveraging high quality data, LIMO \citet{ye_limo_2025} elicits Qwen-2.5-32B-Instruct's reasoning capabilities and matched Qwen-QwQ-32B-Preview's performance. Similarly, \citet{muennighoff_s1_2025} obtains strong results with \(1\)K SFT samples. \citep{carlsson_hyperfitting_2024} proposed Hyperfitting, a method that improves output quality by overfitting small datasets. However, SFT can be not as effective as RL: \citet{li_limr_2025} finds small-scale SFT less effective than RL for smaller models like Qwen-Math-7B. \citet{chu_sft_2025} found SFT tend to memorize but not generalizable to out-of-distribution samples.

\textbf{Theoretical analysis of LLM with RL}: Analyzing LLM is challenging while traditional method failed to predict large neural network's strong generalization ability \citep{azar_general_2024}. Direct Preference Optimization(DPO) propose a practical method to bypass expensive Reinforcement Learning \citep{rafailov_direct_2023}. Massive works based on DPO reveals both empirical and theoretical achievements \citep{ji_towards_2024,ethayarajh_kto_2024}. \citet{ren_learning_2024} established a hypothetical framework using the step-wise decomposition to explain the hallucination in SFT and the squeezing effect appeared in DPO. \citet{koh_understanding_2020} leverage influence function to explain sample contribution. While its purpose is close to our approach, this method focus on minimizing loss on a limited dataset, which is different from reinforcement learning settings. Neural Tangent Kernel(NTK) focus on linearized effect and achieved remarkable success in predicting the behavior of infinite wide networks \citep{jacot_neural_2018,arora_exact_2019}.

\section{Preliminaries} \label{sec:preliminary}

\textbf{GRPO}: We leverage Group Relative Policy Optimization(GRPO) for Reinforcement Learning. Although other RL algorithms have show remarkable performance, such as DAPO \citep{yu_dapo_2025} and REINFORCE++ \citep{hu_reinforce_2025}, we choose GRPO because of its simplicity. GRPO aims to maximize the objective \(\mathcal{J}_{GRPO}(\theta) = \mathbb{E}_{a \sim \pi_{\text{old}}(s), s \sim \mathcal{D}_{\text{train}}}[\frac{1}{G}\sum_{i=1}^{G}(r A_i, \text{clip}(r, 1-\epsilon, 1+\epsilon)A_i) - \beta \mathbb{D}_{KL}(\pi_\theta || \pi_{\text{ref}})]\), where \(r(\pi_\theta, \pi_{\theta_\text{old}}, s, a) = \frac{\pi_\theta(s,a)}{\pi_{\theta_\text{old}}(s,a)}\) is the probability ratio, \(\mathbb{D}_{KL}\) is the estimated KL divergence, \(s\) represents prompt as state, \(a\) represents response as action, \(\pi_\text{old}\) denotes sampling policy, \(\pi_{\text{ref}}\) denotes reference(or initial) policy for KL divergence control.

\textbf{Knight \& Knave dataset(K\&K)}: Knight \& Knave is a logic puzzle that requires identify each person as knight or knave \citep{xie_memorization_2024}. We choose K\&K dataset because it has flexible difficulties and does not rely on prior knowledge(i.e. math theories). By increasing the number of person \(N_{\text{ppl}}\) in one puzzle, it is able to create challenging puzzle for LLM. Solving the puzzle heavily rely on verifying, back-tracking and self-correction abilities. The solution can be synthesized through a solver program, using a template to translate logic expressions into natural language.

\section{Investigating the effectiveness of Small-Scale SFT} \label{sec:method}

A reasonable concern is that small-scale SFT may not substantially change RL behaviors, no matter better or worse, because it only contains a tiny proportion of data compared with pretraining stage. Small-scale SFT may also be unnecessary because the elicited reasoning behavior may be easily found by RL exploration. To address these concerns, we start from observing the effectiveness of small-scale SFT on K\&K and MATH dataset. In the following sections, we use \texttt{-math} and \texttt{-kk} to refer models trained on each dataset. We refer to both datasets if suffix is not specified. Refer to Section \ref{sec:training-details} for training details and Section \ref{sec:exp-settings} for experiment settings.

\subsection{Data Preparation} \label{subsec:data}

On K\&K dataset, we generate two Chain-of-Thought datasets for small-scale SFT: \texttt{short-CoT-kk} contains 1,000 questions and step-by-step solutions synthesized by a solver. \texttt{long-CoT-kk}(846 samples) is generated by distilling DeepSeek-R1 with the same 1K questions and filter incorrect or too long responses. The questions in two datasets have low difficulty as \(N_{ppl} \in \{2, 3\}\). Smaller \(N_{ppl}\) means easier puzzles. For reinforcement learning, we generate 2,000 samples per \(N_{ppl} \in \{3,4,5,6,7\}\) for training, resulting 10K samples. The RL test set contains 500 samples (100 per \(N_{ppl} \in \{4,5,6,7,8\}\)). \(N_{ppl}=8\) samples are only contained in test set to evaluate out-of-distribution generalization.

On MATH dataset, we leave MATH-500 test problems out and further split the rest 12K samples into 11K/900/100 for RL training, SFT training and SFT validation. Similarly, there are two SFT datasets. \texttt{short-CoT-math} uses step-by-step solutions from the original dataset \citep{hendrycks_measuring_2021}. \texttt{long-CoT-math} contains 690 training samples distilled and filtered from DeepSeek-R1.

\subsection{Supervised Fine-tuning and Reinforcement Learning} \label{subsec:SFT-RL}

\begin{figure*}[h]
  \centering
  \subfigure{
  \includegraphics[width=0.45\textwidth]{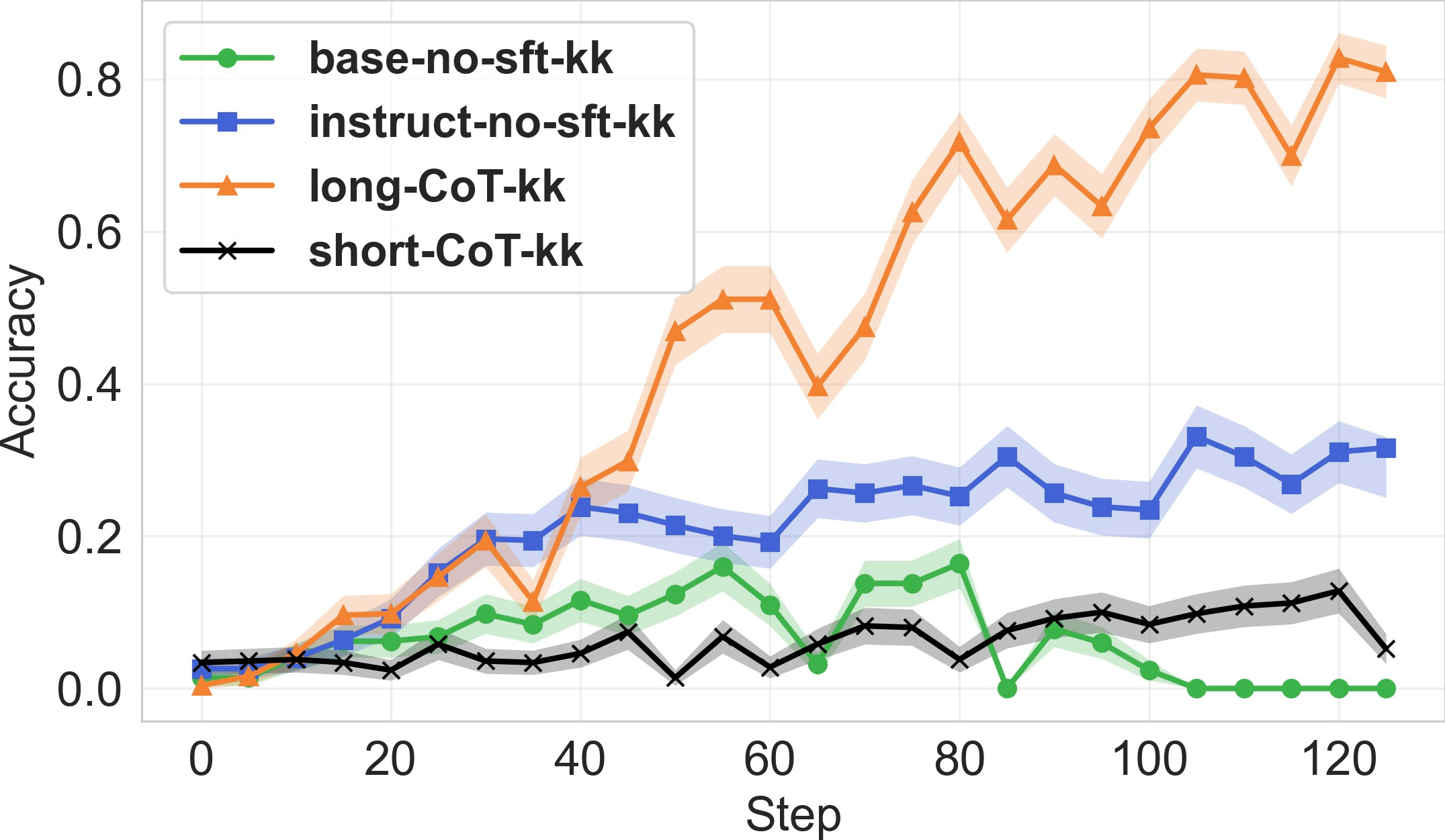}}
  \subfigure{
  \includegraphics[width=0.45\textwidth]{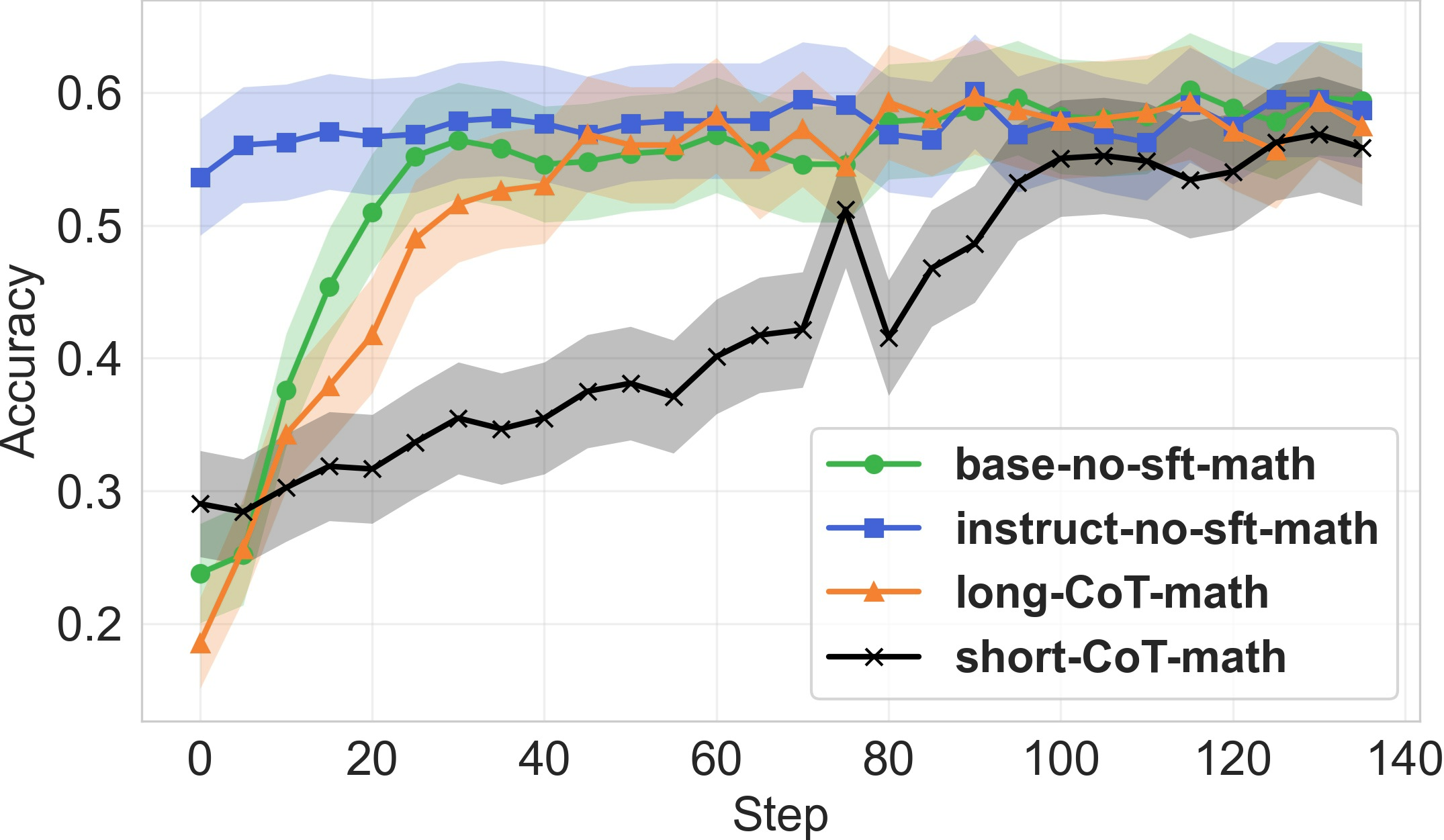}}
  \caption{\textbf{Small-scale SFT affects RL convergence (Left: K\&K dataset, Right: MATH dataset)}.  The RL test accuracy shows that small-scale SFT has substantial influence but lacks of clear pattern. For example, although \texttt{long-CoT-kk} demonstrates superior performance to the other models, \texttt{long-CoT-math} is slightly slower than \texttt{base-no-sft-math}. Shaded area: 95\% Confidence Interval.}
  \label{fig:sft-rl}
\end{figure*}

We fine-tune the Qwen-2.5-1.5B \citep{yang_qwen2_2024-1} Base model on four SFT datasets above. Our first finding is that \textbf{small-scale SFT does not instantly improve performance}. Although distilled \texttt{long-CoT} models learned advanced reasoning behaviors (e.g., "Wait" or "Let me check"), SFT alone shows limited immediate accuracy improvements, as the step 0 accuracy degrades in Figure \ref{fig:sft-rl}. The SFTed model, \texttt{long-CoT-math}, achieves 18.4\% accuracy on MATH test set, which is lower than base model(23.8\%). 

After SFT, we collect four types of initial policies to perform R1-style Reinforcement Learning: The Qwen-2.5-1.5B base model(\texttt{base-no-sft}), Qwen-2.5-1.5B instruct model(\texttt{instruct-no-sft}) and two types of SFTed models(\texttt{short-CoT} \& \texttt{long-CoT}). For each policy, we use the same RL recipe and training set.

Figure \ref{fig:sft-rl} shows several key findings in RL stage. Firstly, \textbf{small-scale SFT substantially affects the RL stage, though the pattern remains unclear}: \texttt{long-CoT-kk} achieves over 80\% accuracy, even surpassed instruct model(\texttt{instruct-no-sft-kk}) with a large margin. \texttt{short-CoT-math} converged much slower. However, DeepSeek-R1 distilled samples are not always the best, as \texttt{long-CoT-math} learns slower than \texttt{base-no-sft-math}. An explanation is that base model has already seen similar pretraining data, so their effects are almost equal. This explanation still follow our findings because we do not assume the affects should be positive. Notably, SFT performance is unreliable for predicting final RL results. \texttt{long-CoT-math} shows lower initial accuracy but converges faster than \texttt{short-CoT-math}, and all K\&K policies begin below 5\% accuracy.

\section{Hypothetical Analysis} \label{sec:theory}

Given the considerable influence of small-scale SFT and the chaotic pattern, we try to evaluate the effectiveness of different type of data from a unified and theoretical perspective. Then, we seek a method to generate data with consistently higher effectiveness. Our analysis is hypothetical, which means it is based on some assumptions to simplify the complex training dynamic. See Subsection \ref{subsec:app-proof-drift} for full proofs and assumptions.

\subsection{Estimate Reward Growth Rate} \label{sebsec:theory-intro}

Intuitively, we need to describe ``how fast test accuracy increases during training''. Let \(a\) denotes a response(action) and \(s\) denotes a prompt(state). With an evaluation dataset \(D_e\) and a policy \(\pi_\theta(a,s)\), the test accuracy can be defined as \(\Psi(\theta)=\mathbb{E}_{a \sim \pi_\theta, s \sim D_e}[r(a,s)]\), where \(r(a,s)\in \{0,1\}\) is a binary reward. Consider optimizing \(r(a,s)\) by a basic REINFORCE \citep{sutton_policy_1999} algorithm and Stochastic Gradient Descent(SGD), policy gradient is computed by aggregating samples as Equation \ref{equ:pg-raw}, where \(N\) is train batch size and \(D_t\) is training set.

\begin{equation} \label{equ:pg-raw}
    -\nabla_\theta L=\frac{1}{N}\sum_{a \sim \pi_\theta, s \sim D_t}[\vec \nabla_\theta \ln \pi_\theta(a,s) r(a,s)]
\end{equation}

When learning rate \(\eta\) is sufficiently small, we can describe the dynamic system by Stochastic Differential Equation(SDE): \( \mathrm{d}\vec \theta=- \vec \nabla_\theta L \mathrm{d}t\). Since test accuracy \(\Psi(\theta(t))\) is a random variable related to \(\vec \theta\), it has mean(drift term \(\mu(t, X)\)) and variance(noise term, \(\sigma(t, X)\)) evolving with time \(t\). We focus on the drift term \(\mu(t, \Psi(\theta))\) because it measures the mean growth rate of test accuracy. We derive the growth rate of test accuracy in Equation \ref{equ:drift-rl}. 

\begin{equation} \label{equ:drift-rl}
\begin{aligned}
    & \mu(t, \Psi(\theta))= \\
    &\underbrace{\underset{\substack{s\sim D_e\\a\sim\pi_\theta}}{\mathbb{E}}[\vec \nabla_\theta \ln \pi_\theta(a,s) r(a,s)]^\top \underset{\substack{s\sim D_t\\a\sim\pi_\theta}}{\mathbb{E}}[\vec \nabla_\theta \ln \pi_\theta(a,s) r(a,s)]}_{\textbf{Positive effect from reward signal}}\\
    & + \underbrace{\frac{1}{2}\frac{\eta^2}{N}\sum_{i=1}^n\vec A_i^\top \nabla^2\Psi(\theta)\vec A_i}_ {\textbf{Negative effect from noise}}
\end{aligned}
\end{equation}

The drift term \(\mu(t, \Psi(\theta))\) contains two terms. The first term is the inner product of policy gradient on training set and test set, which is non-negative when two sets are drawn from the same distribution. It contains the positive effect from reward signal. The second term comes from gradient noise. It reveals how performance deteriorates by random disturbance. For example, on local maximum, the Hessian matrix \(\nabla^2\Psi(\theta)\) is negative definite, so the second term is non-positive. 

With small learning rate and large train batch size, the coefficient \(\eta^2/N\) can be small, we assume that when initial policy is far from converged and hyperparameters are suitable, the positive effect dominates training dynamic as initial steps. Therefore, we only compute the first term \(\hat \mu(t, X)\) as an approximation of \(\mu(t, X)\) in the following analysis.

\begin{definition} \label{def:sample-effect}
    Given evaluation set \(D_e\) and model parameter \(\theta\), we define \textbf{sample effect} \(V(a,s,\theta)\) for each prompt \(s\) and response \(a\) pair as: \\
    \(V(a,s,\theta) = \underset{\substack{s\sim D_e\\a\sim\pi_\theta}}{\mathbb{E}}[\vec \nabla_\theta \ln \pi_\theta(a,s) r(a,s)]^\top \vec \nabla_\theta \ln \pi_\theta(a,s)\)
\end{definition}

We define the \textbf{sample effect} \(V(a,s,\theta)\) in Definition \ref{def:sample-effect}. Intuitively, it describes the contribution of a sample's gradient computed by cross-entropy loss. By introducing sample effect, the approximated growth rate \(\hat \mu(t, X)\) can be written as a weighted average of sample effect in Equation \ref{equ:mu-approx-rl}.

\begin{equation} \label{equ:mu-approx-rl}
    \hat{\mu}_{\text{RL}}(t, \Psi(\theta)) = \mathbb{E}_{s\sim D_t, a\sim\pi_\theta}[V(a,s,\theta) r(a,s)]
\end{equation}

The definition of sample effect is similar with influence function, which added a Hessian matrix term \citep{koh_understanding_2020}. The main difference is that sample effect focus on training efficiency of initial steps, while influence function aims to predict the loss near convergence. In R1-style Reinforcement Learning, policy can be not converged even after a thousand steps, hence sample effect may offer a more helpful insight. 

\subsection{Optimal Policy for Filtered Distillation} \label{subsec:theory-opt-target}

We want to find the optimal target policy to maximize the effectiveness of distilling a given policy and the connections between optimal policy and sample effect. 
Consider generate samples from a target policy \(\pi_\theta^\dagger(a,s)\) on training set and distill \(\pi_\theta(a,s)\) with only correct samples. This can be viewed as distilling from \(\hat{\pi}_\theta^\dagger(a,s)=\frac{\pi_\theta^\dagger(a,s)}{p_\theta^\dagger(s)}r(a,s)\) where \(p_\theta^\dagger(s)=\mathbb{E}_{s\sim D_t, a\sim \pi_\theta^\dagger}[r(a,s)]\) is the accuracy for each problem \(s\).

\begin{equation} \label{equ:mu-approx-sft}
    \hat{\mu}_{\text{SFT}}(t, \Psi(\theta)) = \mathbb{E}_{s\sim D_t, a\sim \hat{\pi}_\theta^\dagger}[V(a,s,\theta) r(a,s)]
\end{equation}

Using sample effect, we compute the growth rate of SFT in Equation \ref{equ:mu-approx-sft}. Then, we derive how fast SFT improves than RL (Equation \ref{equ:drift-delta}).

\begin{align} \label{equ:drift-delta}
    \Delta \hat{\mu}(t, \Psi(\theta)) = & \hat{\mu}_{\text{SFT}}(t, \Psi(\theta)) - \hat{\mu}_{\text{RL}}(t, \Psi(\theta)) \\
    = & \underset{\substack{s\sim D_t\\a\sim\pi_\theta}}{\mathbb{E}}[V(a,s,\theta) (\frac{\pi_\theta^\dagger(a,s)}{p_\theta^\dagger(s) \pi_\theta(a,s)} - 1) r(a,s)]
\end{align}

Given an initial policy \(\pi_\theta\) and a dataset, an optimal target policy \(\pi_\theta^*\) maximizes delta speed \(\Delta \hat{\mu}(t, \Psi(\theta))\), where \(p_\theta^*(s)=\mathbb{E}_{a\sim \pi_\theta^*}[r(a,s)|s]\) is a constraint on \(\pi_\theta\). We wonder the relationship between optimal policy and sample effect, as well as \(p_\theta^*(s)\). Therefore, we compute the optimal distribution by maximizing \(\Delta \hat{\mu}(t, \Psi(\theta))\) with a KL divergence constraint, which induces a coefficient \(\beta\).

\begin{theorem} \label{theorem:opt-sft-policy}
Let \(\pi_\theta^*\) to be the optimal policy for maximizing \(\mathbb{E}_{s\sim D_t, a\sim\pi_\theta}[V(a,s,\theta) (\frac{\pi_\theta^*(a,s)}{p_\theta^*(s) \pi_\theta(a,s)} - 1) r(a,s)] - \beta \mathbb{D}_{KL}[\pi_\theta^*||\pi_\theta]\) with \(\beta > 0\) and \(\mathbb{E}_{a\sim \pi_\theta^*}[r(a,s)|s] = p_\theta^*(s) \in (0,1]\). For any \(a_1, a_2\) and \(s\in \{D_t\}\) that satisfies \(r(a_1,s)=r(a_2,s)=1\), we have: \(\ln{\frac{\pi_\theta^*(a_1,s)}{\pi_\theta^*(a_2,s)}} =\frac{1}{\beta p_\theta^*(s)}(V(a_1,s,\theta)-V(a_2,s,\theta)) +  \ln\frac{\pi_\theta(a_1,s)}{\pi_\theta(a_2,s)}\)
\end{theorem}

\begin{corollary} \label{cor:opt-sft-p}
For any optimal policy \(\pi_\theta^*\) with \(p_\theta^*(s) \in (0, 1]\) and \(\beta > 0\), we can always find another \(\beta'\) and \(p_\theta'(s)\) which satisfies \(\beta p_\theta^*(s)  = \beta' p_\theta'(s) \) to make the filtered optimal policy \(\hat \pi_\theta^*\) unchanged.
\end{corollary}

In Theorem \ref{theorem:opt-sft-policy}, we proved that answers with higher \(V(a,s,\theta)\) tend to gain more probability mass in optimal target policy. When \(V(a_1,s,\theta)>V(a_2,s,\theta)\) and \(\beta\) is small enough, the output probability of \(a_1\) will be greater than \(a_2\). Therefore, samples with higher effect will obtain more probability mass. But does the test accuracy matter? In Corollary \ref{cor:opt-sft-p}, we proved that simply adjusting accuracy is identical to choosing a different \(\beta\), which does not improve SFT effectiveness. 

\subsection{Reinforcement Learning Improves Output Sample Effect} \label{subsec:theory-rl-effect}

While Theorem \ref{theorem:opt-sft-policy} suggests that distillation benefits from a target policy with high output effects, it does not tell how to find the optimal policy. Motivated by the similarity between initial policy and RL-trained policy, we investigate if RL-trained policy is a potential answer. In Definition \ref{def:dataset-effect}, we introduce \textbf{dataset effect}, which is a weighted average of sample effect in training set \(D_t\). Note that we sample from filtered policy \(\hat{\pi}_\theta^\dagger\) but not the original policy \(\pi_\theta^\dagger\).

\begin{definition} \label{def:dataset-effect}
Let \(\pi_\theta^\dagger\) to be any policy and \(\hat{\pi}_\theta^\dagger\) to be the filtered target policy. Given a training set \(D_t\), we define the \textbf{dataset effect} as: \(V(\theta, \theta^\dagger, D_t)=\mathbb{E}_{s\sim D_t, a\sim \hat{\pi}_\theta^\dagger}[V(a,s,\theta)r(a,s)]\). 
\end{definition}

\begin{theorem} \label{theorem:rl-effect}
Let \(D_t\) denotes the training set and \(D_e\) denotes the evaluation set. \(\hat{\mu}(t, V(\theta, \theta^*, D_t))\) denotes the approximated growth rate of dataset effect in RL. \(\hat{\mu}(t, \Psi(\theta))\) denotes the approximated growth rate of reward. When \(D_t=D_e\) and \(\theta^*=\theta\), we have \(\hat{\mu}(t, V(\theta, \theta^*, D_t)) \geq \hat \mu^2(t, \Psi(\theta))\).
\end{theorem}

In Theorem \ref{theorem:rl-effect}, we assume that RL is applied on \(\pi_\theta\). If we observed a positive growth of training accuracy \(\hat{\mu}(t, \Psi(\theta))\), a positive increment of filtered dataset effect \(V(\theta, \theta^*, D_t)\) is guaranteed at least at initial steps. This analysis reveals great potential of distillation from RL-trained policy. Notably, the increment is not simply come from correctness filtering because \(\hat{\mu}(t, V(\theta, \theta^*, D_t)) > 0\) indicates the growth rate based on \(\mathbb{E}_{s\sim D_t, a\sim \hat{\pi}_\theta^*}[V(a,s,\theta)r(a,s)]\), which is already filtered. 

However, the theoretical guarantee holds only under some limitations. Firstly, \(D_t=D_e\) requires collecting questions in the same distribution as test set. Secondly, \(\theta^*=\theta\) means both source and target models share the same(or similar) pretrained model. Thirdly, our analysis ignored the non-linear effect when \(\theta^*\) is gradually diverged from \(\theta\). If we perform RL for \(\theta^*\) with more steps, the improvement of dataset effect may decrease to zero. Therefore, we perform empirical study to examine the boundary of our analysis in following sections.

\section{Re-distillation from RL-trained Models} \label{sec:re-distill}

\begin{figure*}[t]
  \centering
  \subfigure{
  \includegraphics[width=0.45\textwidth]{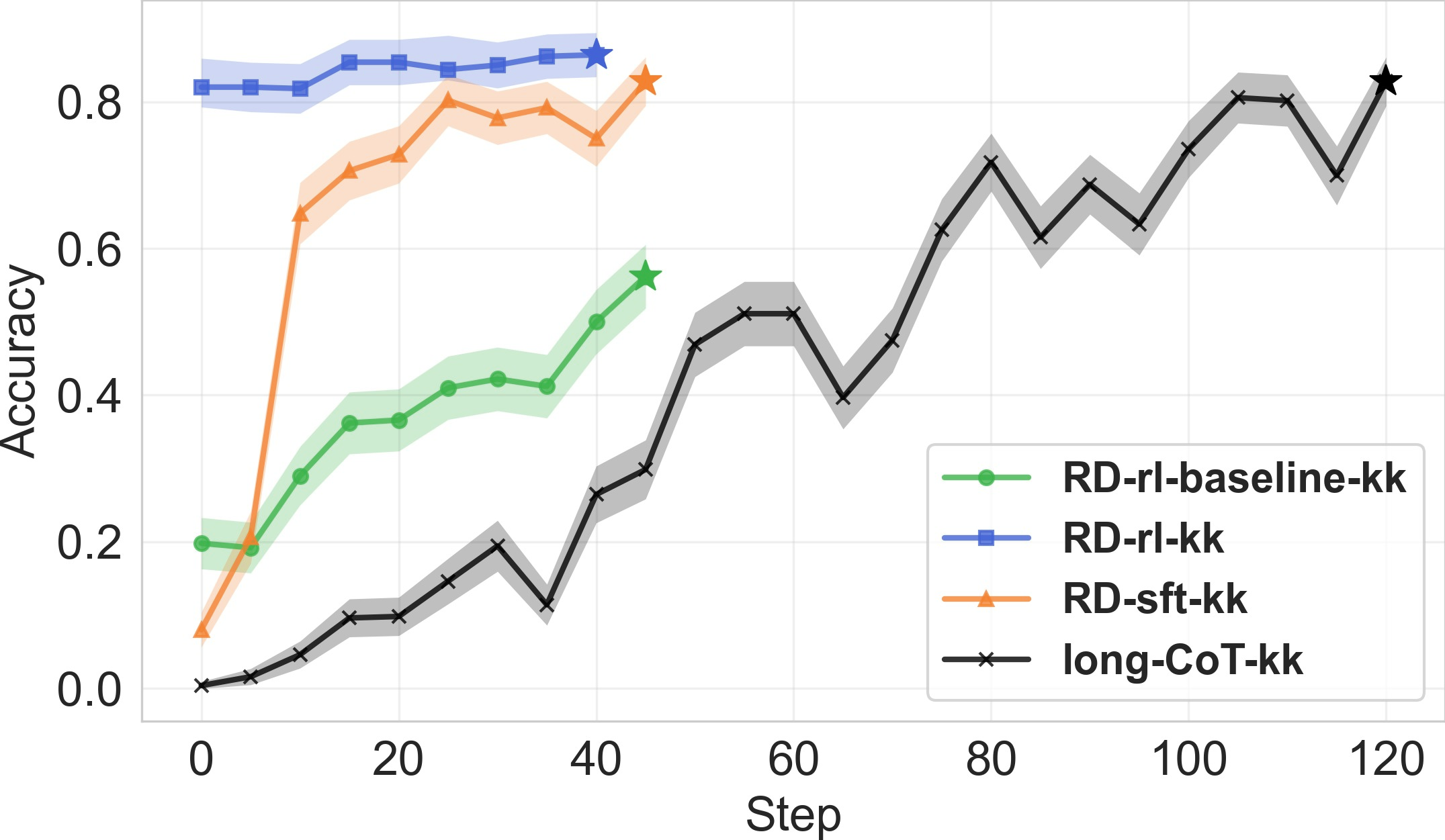}}
  \subfigure{
  \includegraphics[width=0.45\textwidth]{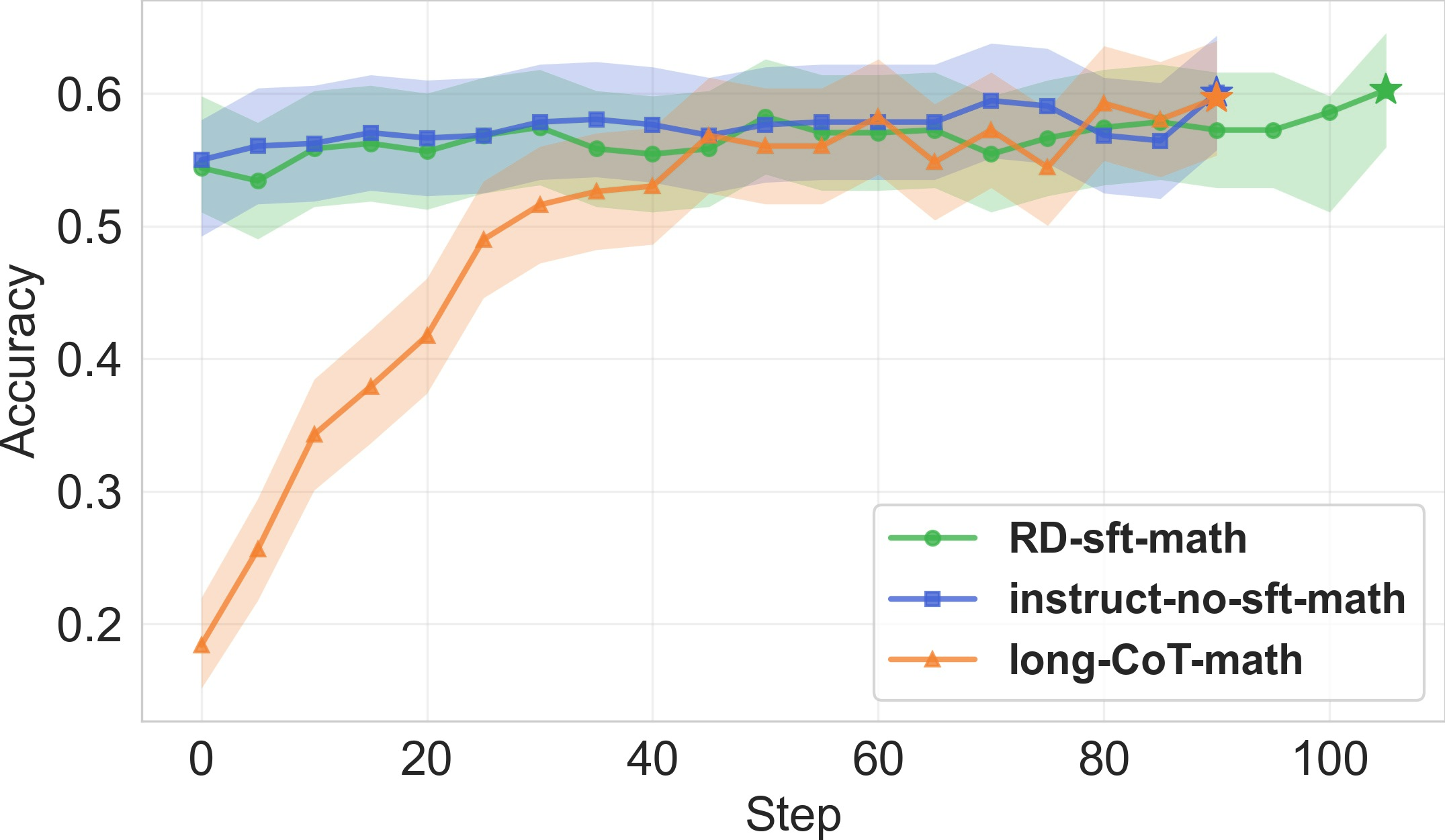}}
  \caption{\textbf{Re-distillation matches RL performance through SFT alone. (Left: K\&K dataset, Right: MATH dataset)} At step 0, \texttt{RD-sft-math} matches Qwen-2.5-Instruct by only SFT on 496 samples. \texttt{RD-rl-kk} approaches optimal performance without RL. Re-distillation also boost RL efficiency, as \texttt{RD-sft-kk} converges significantly faster than \texttt{long-CoT-kk}. Shaded area: 95\% Confidence Interval.} 
  \label{fig:RD-acc}
\end{figure*}

\begin{figure*}[h]
  \centering
  \subfigure{
  \includegraphics[width=0.34\textwidth]{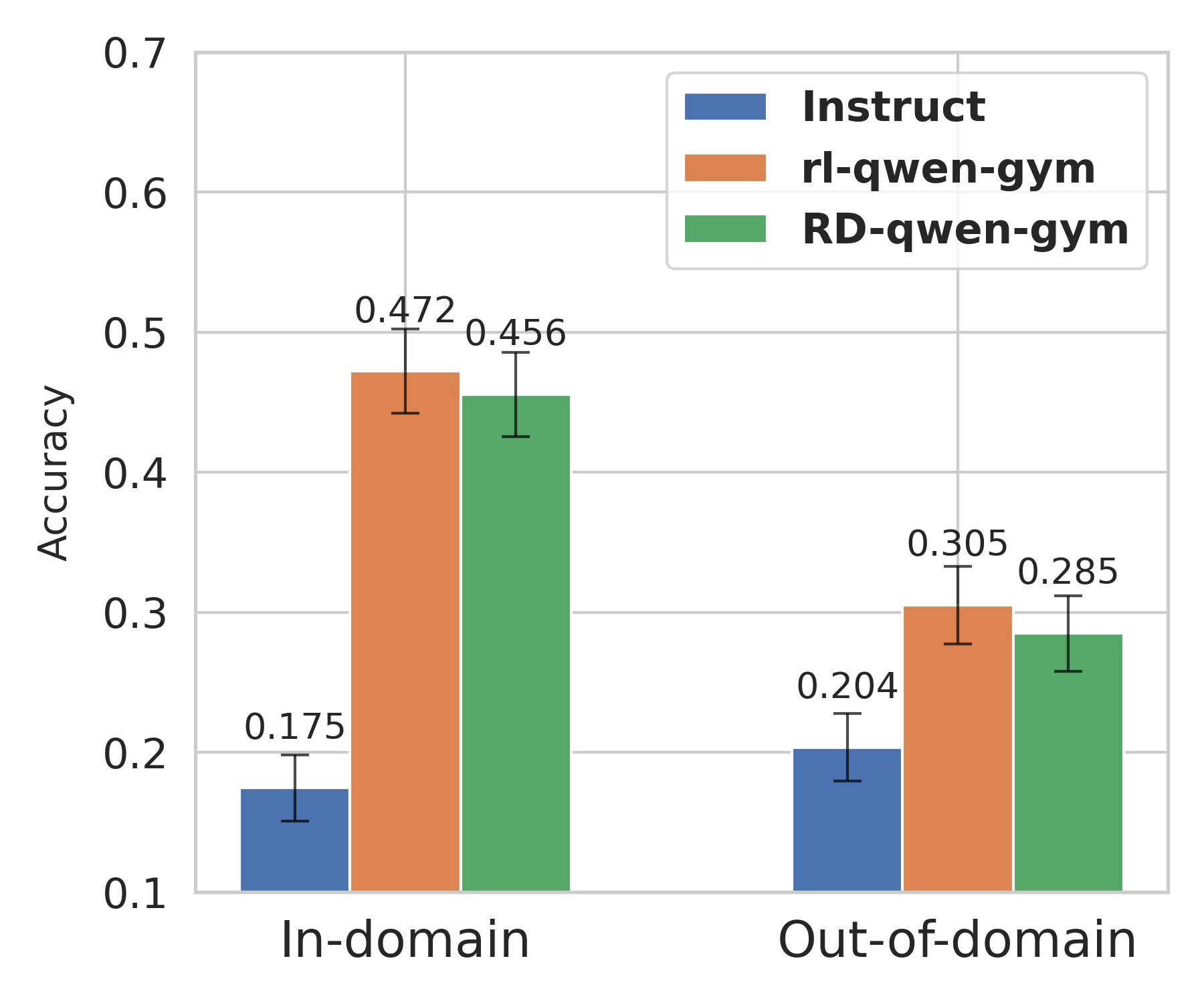}}
  \subfigure{
  \includegraphics[width=0.56\textwidth]{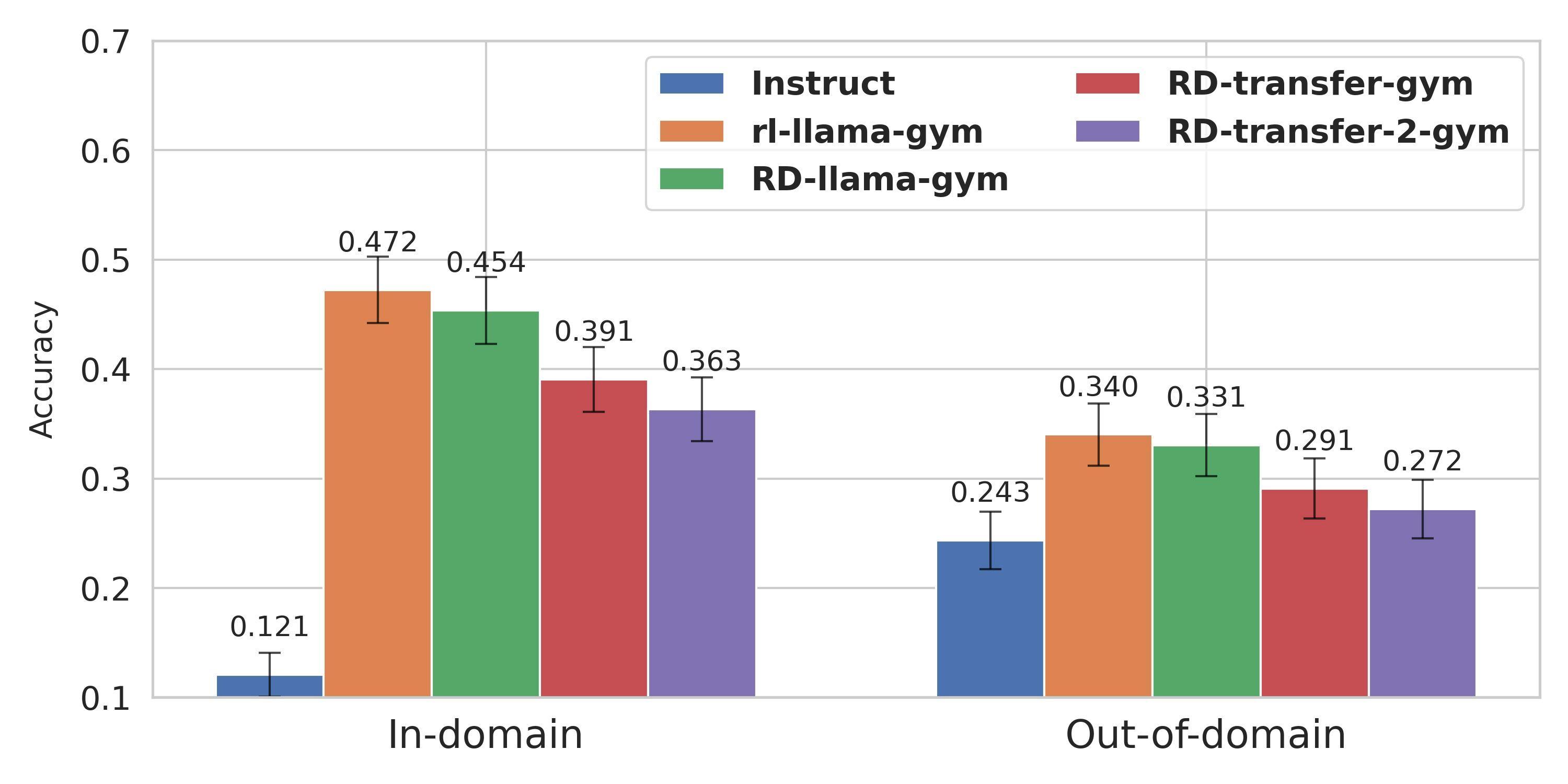}}
  \caption{\textbf{Re-distillation is generalizable on variant tasks and models (Left: Qwen, Right: Llama)}: RL from Instruct models gains substantial improvement(Instruct to \texttt{rl-}). The re-distilled model from RL-trained policy retains most performance improvement(\texttt{RD-} vs \texttt{inst-}). However, when Llama re-distilling from Qwen model, the effectiveness decreases as \texttt{RD-transfer-gym} and \texttt{RD-transfer-2-gym}. Error bar: 95\% Confidence Interval.} 
  \label{fig:RD-gym}
\end{figure*}

Inspired by the above analysis, we use a re-distillation approach to create dataset with high sample effects. This method involves distilling the RL-trained policy back to initial model. We validate this approach on three datasets(K\&K, MATH and REASONING GYM) and two models(Qwen-2.5-1.5B and Llama-3.2-3B). At the end of this section, we provide an application case to show how re-distillation merges RL-trained policies without RL. We provide training details in Section \ref{sec:training-details} and experiment settings in Section \ref{sec:exp-settings}.

\subsection{Two Simple Cases: K\&K and MATH}

On K\&K dataset, we generate responses using the 125-step \texttt{long-CoT-kk} checkpoint with correctness and length filtering. 
There are two types of re-distillation datasets. Here we use \texttt{RD-rl-} and \texttt{RD-sft-} to indicate the source dataset which we re-distill from. \texttt{RD-sft-kk}(689 samples) uses the same questions(\(N_{ppl}=\{2,3\}\)) as SFT dataset in \texttt{long-CoT-kk} and \texttt{short-CoT-kk}. So it is fair to compare the effectiveness of different cold-start SFT data.
\texttt{RD-rl-kk} consists of randomly selected 1K questions directly from RL training set of \texttt{long-CoT-kk}. To check if \texttt{RD-rl-kk} improves by simply changing the dataset, we use the same 1K questions to prompt DeepSeek-R1, then filter responses to create \texttt{RD-rl-baseline-kk}. On MATH dataset, we create \texttt{RD-sft-math}(496 samples) dataset from \texttt{long-CoT-math} checkpoint. Questions are as same as \texttt{long-CoT-math} and responses are generated by checkpoint at 50 step. This dataset has no difficulty shift compared with test set. As all three requirements in Theorem \ref{theorem:rl-effect} are nearly satisfied, we expect \texttt{RD-sft-math} and \texttt{RD-rl-kk} show higher SFT effectiveness.

As shown in Figure \ref{fig:RD-acc}, \textbf{re-distilled models demonstrate matched accuracy and accelerated convergence}. Comparing re-distilled policy with RL-trained policy, re-distillation enables a direct performance matching by only SFT: At step 0, \texttt{RD-sft-math} attains similar test accuracy as \texttt{instruct-no-sft-math}(54.4\% vs 55.0\%). It also matched its source policy, 50-step \texttt{long-CoT-math} checkpoint(54.4\% vs 56.0\%). \texttt{RD-rl-kk} is close to \texttt{long-CoT-kk} through SFT alone (78.8\% vs 82.0\%), which is surprising because \texttt{base-no-sft-kk} has near zero accuracy and such great improvement is achieved by only 1K samples. 

Re-distilled models also converged faster. \texttt{RD-sft-kk} and \texttt{long-CoT-kk} use the same questions in SFT stage but \texttt{RD-sft-kk} reaches 80.2\% test accuracy within 25 steps, showing \(5\times\) higher efficiency than \texttt{long-CoT-kk}. The performance gap between \texttt{RD-sft-kk} and \texttt{RD-rl-kk} may caused by distribution shift, as \texttt{RD-sft-kk} uses simpler data which does not align with RL training set.

Does the improvement of \texttt{RD-rl-kk} come from simply changing the dataset? Comparing \texttt{RD-rl-kk} with \texttt{RD-rl-baseline-kk}, re-distilled policy still shows significant improvement, which indicates the re-distilled responses are more efficient.

\subsection{A Challenging Task: REASONING GYM}

While re-distillation is effective on the above two datasets, we wonder if it is generalizable on more tasks and models. Therefore, we design more challenging experiments on REASONING GYM\citep{stojanovski2025reasoninggym}, which provide 100+ verifiable tasks in 10 categories. For creating RL dataset, we select 20 tasks as in-domain tasks and another 20 tasks as out-of-domain tasks. Out-of-domain tasks will only appear in test set. Our selected tasks covered all categories and have various difficulties. Model needs to learn 20 tasks simultaneously from the synthesized dataset. 

We run GRPO on Qwen-2.5-1.5B Instruct (\texttt{rl-qwen-gym}) and Llama-3.2-3B Instruct (\texttt{rl-llama-gym}) without SFT. After RL for 100 steps, we generate \(1K\) samples from replay buffer and create two re-distillation datasets(\texttt{RD-qwen-gym} and \texttt{RD-llama-gym}), which contain only in-domain tasks. We re-distill initial RL policies, the Instruct models, to observe performance improvement. 

As shown in Figure \ref{fig:RD-gym}, both Instruct models have gained substantial improvement during RL. Two re-distilled models' average improvement in in-domain tasks are greater than out-of-domain tasks(32.4\% vs 9.9\%). We found that \textbf{re-distilled models retain most improvements by SFT with \(1K\) samples}, only lose 1.7\% accuracy on in-domain tasks and 1.5\% on out-of-domain tasks on average. These results are surprising because over \(200K\) responses are generated during RL, but the improvement is recovered by about 50 samples for each task.

According to Theorem \ref{theorem:rl-effect}, the effectiveness of re-distillation may diminish when source policy is different from target policy. But there are also other possible scenarios. For example, target policy can still obtain benefit from source policy once target model is larger because of higher model capacity. Another model agnostic hypothesis is that target policies can reach the same accuracy once they trained on the same SFT dataset. We address these concerns by introducing \texttt{RD-transfer-gym}, a model SFTed from Llama-3.2-3B Instruct on \texttt{RD-qwen-gym} dataset. We only change the prompt template of \texttt{RD-qwen-gym} to llama style. In Figure \ref{fig:RD-gym} Right, \texttt{RD-transfer-gym} failed to reach the in-domain accuracy of either re-distilled models. In 16 in-domain tasks trained on \texttt{RD-qwen-gym} dataset, the average accuracy of \texttt{RD-transfer-gym} is lower than \texttt{RD-qwen-gym}(47.5\% vs 55.4\%), despite they have the same SFT dataset. \texttt{RD-transfer-2-gym} collects responses in \texttt{RD-qwen-gym} according to the questions proportion of \texttt{rl-llama-gym} also failed to match \texttt{rl-llama-gym}. See experiment settings in Section \ref{sec:exp-settings} for details. This experiment reveals \textbf{the model-specific property of re-distillation}, which aligned with our analysis in Section \ref{sec:theory}.

\subsection{An application: Merging RL Policies} \label{subsec:merge}

In this section, we demonstrate how to leverage re-distillation to merge two RL-trained polices. A large scale reinforcement learning usually contain multiple goals, such as improving reasoning ability and enhancing safety. Balancing them during RL is sometimes difficult. By compressing RL process into SFT, it is easier to merge RL-trained policies or adjust task weights.

\begin{table}[h]
\centering
\caption{Test accuracies. RG: REASONING GYM, ID: in-domain tasks, OD: out-of-domain tasks}
\label{tab:app-combined}
\begin{tabular}{c c c c}
\toprule
Model & RG(ID) & RG(OOD) & K\&K \\
\midrule
\texttt{RD-rl-kk} & 14.9\% & 24.5\% & 82.0\% \\
\texttt{RD-qwen-gym} & 45.6\% & 28.5\% & 4.4\% \\
\texttt{RD-combined} & 43.0\% & 26.5\% & 77.6\% \\
\bottomrule

\end{tabular}
\end{table}

We combine \texttt{RD-rl-kk} dataset and \texttt{RD-qwen-gym} dataset to create \texttt{RD-combined} dataset. As shown in Table \ref{tab:app-combined}, the combined policy reveals significant improvement in RG(ID) and K\&K, gaining abilities from both datasets. As SFT uses less compute than RL, this approach saves a lot of time in re-training RL to move along the Pareto frontier.

\section{Discussion and Conclusion}

\textbf{Limitations}: This study has several limitations. Firstly, linearized sample effect may not be able to fully explain the non-linear effect in SFT and RL. Re-distilled model still has minor but not zero performance gap compared with RL-trained policy. Secondly, re-distillation is a post-hoc method which requires RL-trained policy. It remains a problem to select effective samples before SFT or RL. Thirdly, our experiments are based on small LLMs(1.5B and 3B). The effectiveness of re-distillation is unknown on large LLMs or large scale RL with more steps.

\textbf{Conclusion}: This work explained the success of R1-style RL. We investigate the mechanisms behind R1-style Reinforcement Learning (RL) and demonstrate that small-scale Supervised Fine-Tuning (SFT) can achieve comparable performance to RL. Our method is generalizable on different datasets and models. Theoretical and empirical analysis reveal that sample effect is a key indicator of model improvement, explaining why RL-trained policies can generate effective distillation data.

\section{Acknowledgement}

This work is funded by Noncommunicable Chronic Diseases-National Science and Technology Major Project(Grant No. 2024ZD0522702)

\bibliography{DeepReasoning}


\newpage

\appendix

\onecolumn

\section{Experiment Settings} \label{sec:exp-settings}

We list all Supervised Fine-Tuning, Reinforce Learning and Re-distillation experiments as well as other baselines as follows:

\textbf{K\&K Experiments}:

In RL stage, the training set contains 10K samples(\(N_{ppl} \in \{3,4,5,6,7\}\)). We use 500 held out samples(\(N_{ppl} \in \{4,5,6,7,8\}\)) as RL test set. All models on K\&K dataset have the same RL recipe(i.e. hyperparameters, RL dataset, etc.). 

\begin{itemize}
    \item \texttt{short-CoT-kk} 1K simple questions(\(N_{ppl} \in \{2, 3\}\)) synthesized by program. Step-by-step answers are generated by a solver program. In SFT stage, we fine-tune Qwen-2.5-1.5B Base with this dataset. 
    \item \texttt{long-CoT-kk} The solutions for SFT are generated by prompting DeepSeek-R1. We keep correct responses within max response length to build SFT dataset. Other parts are as same as \texttt{short-CoT-kk}.
    \item \texttt{base-no-sft-kk} We apply RL on Qwen-2.5-1.5B Base model without SFT stage. 
    \item \texttt{instruct-no-sft-kk} We apply RL on Qwen-2.5-1.5B Instruct model without SFT stage. 
    \item \texttt{RD-sft-kk} By prompting the  125-step RL checkpoint of \texttt{long-CoT-kk}, we generate and filter correct responses on the SFT question set of \texttt{long-CoT-kk}. Other parts are as same as \texttt{long-CoT-kk}.
    \item \texttt{RD-rl-kk} We randomly select 1K correct samples from the RL replay buffer of \texttt{long-CoT-kk} near 125 step. In re-distillation, we fine-tune Qwen-2.5-1.5B Base model with these 1K samples and perform an extra RL stage.
    \item \texttt{RD-rl-baseline-kk} In SFT stage, we use the same 1K questions as \texttt{RD-rl-kk} and generate solutions by prompting DeepSeek-R1, then perform RL. This experiment is aimed to compare with \texttt{RD-rl-kk} as a baseline.
\end{itemize}

\textbf{MATH Experiments}: 

We split 12K MATH dataset into 11K for RL training, 900 for SFT training and 100 for SFT validation. The test set contains 500 held out questions. All models on MATH dataset have the same RL recipe(i.e. same hyperparameters, dataset, etc.). 

\begin{itemize}
    \item \texttt{short-CoT-math} Solutions for this experiment is gathered from original dataset. Each question has a brief human-written solutions. We apply an SFT stage following a RL stage.
    \item \texttt{long-CoT-math} We prompt DeepSeek-R1 to generate solutions and only keep correct responses within max length. The SFT questions and other partsare as same as \texttt{short-CoT-math}.
    \item \texttt{instruct-no-sft-math} Directly run RL on Qwen-2.5-1.5B Instruct without SFT stage.
    \item \texttt{based-no-sft-math} Directly run RL on Qwen-2.5-1.5B Base without SFT stage.
    \item \texttt{RD-sft-math} We use the same SFT questions as \texttt{short-CoT-math} and prompt \texttt{long-CoT-math} checkpoint to generate solutions, then filter to keep only correct answers. Note that these SFT samples have no distribution shift with MATH-500 test set, so it is unnecessary to generate \texttt{RD-rl-math}.
\end{itemize}

\textbf{REASONING GYM Experiments}:

The RL dataset contains 15K synthesized questions(20 in-domain tasks) in training set and 2K samples in test set(20 in-domain + 20 out-of-domain tasks, 50 questions per task). The choice of tasks is based on Table 4 (\citet{stojanovski2025reasoninggym} appendix A.1). We simply select the first 2 tasks for each category as in-domain task and subsequent 2 tasks as out-of-domain tasks. Some tasks are skipped because they either produce errors during generation or can not adjust difficulties. We also change the generation parameters of included tasks to reduce difficulties. The reward functions are provided by dataset repository. Although the reward is a continuous score from 0 to 1, most scores are distributed near 0 or 1. Therefore, we define a response is correct if its score is above 0.95. 

\begin{itemize}
    \item \texttt{rl-qwen-gym} We train Qwen-2.5-1.5B Instruct for 100 steps on RL dataset described above. This experiment is aimed to verify re-distillation, so we did not include an SFT stage before RL. 
    \item \texttt{RD-qwen-gym} The re-distillation dataset is gathered from \texttt{rl-qwen-gym} RL replay buffer. Unlike \texttt{RD-sft-math}, we did not generate responses from checkpoint because some tasks have low correct rate so collecting enough correct samples can be expensive. Instead, we manually assign sample proportion for each task to select data. This approach is aimed to prevent any task occupying too many samples. While the change of proportion affects SFT result, it does not contradict our theoretical assumption because the responses are randomly picked with correctness filter.
    \item \texttt{rl-llama-gym} We replace initial policy from Qwen-2.5-1.5B Instruct to Llama-3.2-3B Instruct and keep other parts as same as \texttt{rl-qwen-gym}.
    \item \texttt{RD-llama-gym} The re-distillation dataset is gathered from \texttt{rl-llama-gym} RL replay buffer. We use a different proportion described below. 
    \item \texttt{RD-transfer-gym} We change the prompt template of \texttt{RD-qwen-gym} dataset and SFT Llama-3.2-1.5B Instruct. We ensure both \texttt{RD-transfer-gym} and \texttt{RD-transfer-2-gym} trained with enough max token length so that no responses will be clipped. The test response length also enlarged to 4096 tokens.
    \item \texttt{RD-transfer-2-gym} While the responses of \texttt{RD-transfer-gym} are sampled from \texttt{RD-qwen-gym}, it uses task proportions of Qwen model, which is different from \texttt{RD-llama-gym}. This experiment randomly select responses from \texttt{RD-qwen-gym} replay buffer, but using Llama proportions. Each task has the similar number of samples compared with \texttt{RD-llama-gym}. Responses and prompts are adjusted for Llama templates
    \item \texttt{RD-combined} We combine the re-distillation dataset of \texttt{RD-rl-kk} and \texttt{RD-qwen-gym} to improve performance on both datasets.
\end{itemize}

Task proportions for creating \texttt{RD-qwen-gym} and \texttt{RD-transfer-gym}: basic\_arithmetic (11.0\%), bitwise\_arithmetic (5.0\%), arc\_1d (2.0\%), polynomial\_equations (8.0\%), graph\_color (8.0\%), chain\_sum (4.0\%), base\_conversion (9.0\%), course\_schedule (10.0\%), pool\_matrix (5.0\%), boxnet (2.5\%),  countdown (0.5\%), aiw (17.0\%), binary\_alternation (4.0\%), ab (2.0\%), modulo\_grid (9.0\%), advanced\_geometry (3.0\%).

Task proportions for creating \texttt{RD-llama-gym} and \texttt{RD-transfer-2-gym}: basic\_arithmetic (4.0\%), bitwise\_arithmetic (2.0\%), arc\_1d (0.5\%), polynomial\_equations (7.0\%), graph\_color (10.0\%), intermediate\_integration (10.5\%), chain\_sum (5.0\%), base\_conversion (10.0\%), course\_schedule (12.0\%), pool\_matrix (3.0\%), boxnet (10.0\%), aiw (11.0\%), binary\_alternation (4.5\%), ab (2.0\%), modulo\_grid (9.0\%).

\textbf{Hyperparameter Tuning Experiments}: 

\begin{itemize}
    \item In K\&K experiments, we found setting \(\beta_1=0.5\) in Adam optimizer enables better RL convergence. We set \(\beta_1=0.9\) for comparison. See Figure \ref{fig:hyper-beta1} for results. 
    \item To see the effect of exploration temperature, in Subsection \ref{subsec:exploration}, we set rollout temperature as 1.2 in \texttt{long-CoT-temp1.2-math} and compare results with \texttt{long-CoT-math}, which uses temperature 0.7. 
\end{itemize}

\textbf{More SFT data}: 

We investigate how more SFT data/epochs affect performance in Subsection \ref{subsec:more-sft}.

\begin{itemize}
    \item  In \texttt{hyperfit-math}, we use HyperFitting to fit SFT data in \texttt{long-CoT-math} with 15 epochs then perform RL. 

    \item In \texttt{distill-50k-math}, we SFT Qwen-2.5-1.5B Base with 50K data filtered from OpenR1-Math 220K and perform RL.
\end{itemize}

\newpage
\section{Experiments Results} \label{sec:exp}

We report the main results for all experiments in Table \ref{tab:kk-acc} and Table \ref{tab:math-acc}. Detailed statistics are depicted in Section \ref{sec:appendix-stats}.

\begin{table*}[htbp]
\centering
\caption{Test accuracy in K\&K dataset}
\label{tab:kk-acc}
\begin{tabular}{>{\raggedright\arraybackslash}p{9cm}c c c c c c}
\toprule
Model/$N_{ppl}$ & 4 & 5 & 6 & 7 & 8 & Average \\
\midrule
Deepseek-R1 & 0.99 & 0.99 & 0.95 & 0.96 & 0.95 & 0.968 \\
Logic-RL-7B\(^\ddag\) & 0.94 & 0.92 & 0.91 & 0.80 & 0.67 & 0.848 \\
Deepseek-R1-Qwen-Distill-32B & 0.95 & 0.91 & 0.85 & 0.87 & 0.79 & 0.874 \\
Deepseek-V3-0324 & 0.85 & 0.84 & 0.86 & 0.81 & 0.68 & 0.808 \\
Deepseek-R1-Qwen-Distill-14B & 0.92 & 0.84 & 0.81 & 0.82 & 0.71 & 0.820 \\
Deepseek-R1-Qwen-Distill-7B & 0.32 & 0.28 & 0.12 & 0.14 & 0.02 & 0.176 \\
Qwen-2.5-32B-Instruct & 0.41 & 0.38 & 0.19 & 0.15 & 0.15 & 0.256 \\
Qwen-2.5-7B-Instruct & 0.18 & 0.17 & 0.01 & 0.05 & 0.01 & 0.084 \\
\midrule
Qwen-2.5-1.5B-Instruct(\texttt{instruct-no-sft-kk}) & 0.03 & 0.05 & 0 & 0.01 & 0 & 0.018 \\
\qquad w/ long CoT RL(\texttt{instruct-no-sft-kk}) & 0.49 & 0.42 & 0.29 & 0.18 & 0.20 & 0.316 \\
Qwen-2.5-1.5B-Base(\texttt{base-no-sft-kk}) & 0.03 & 0.01 & 0.01 & 0 & 0 & 0.010 \\
\qquad w/ long CoT SFT(\texttt{long-CoT-kk}) & 0.01 & 0 & 0 & 0 & 0 & 0.002 \\
\qquad w/ long CoT SFT + RL(\texttt{long-CoT-kk}) & 0.91 & 0.92 & 0.84 & 0.78 & 0.65 & 0.820 \\
\qquad w/ DeepSeek-R1 distillation\(^\dagger\)(\texttt{RD-rl-baseline-kk}) & 0.33 & 0.26 & 0.15 & 0.12 & 0.09 & 0.190 \\
\qquad \textbf{w/ Re-distillation}\(^\dagger\)(\texttt{RD-rl-kk}) & 0.94 & 0.90 & 0.83 & 0.74 & 0.69 & 0.820 \\
\qquad \textbf{w/ Re-distillation + RL}\(^\dagger\)(\texttt{RD-rl-kk}) & 0.97 & 0.91 & 0.85 & 0.80 & 0.69 & 0.844 \\
\bottomrule
\multicolumn{7}{l}{\(\ddag\): We collect this result from paper. We share identical test set with Logic-RL.} \\
\multicolumn{7}{l}{\(\dagger\): We use $L_{max}=8192$ in these models' evaluation.} \\

\end{tabular}
\end{table*}

\begin{table*}[htbp]
\centering
\caption{Test accuracy in MATH-500 dataset}
\label{tab:math-acc}
\begin{tabular}{>{\raggedright\arraybackslash}p{5cm}c c c c}
\toprule
Model &  SFT samples &  \makecell[c]{pass@1 \\(0 step)} & \makecell[c]{pass@1 \\w/ SFT+RL(100 step)} & \makecell[c]{pass@1 \\w/ SFT+RL(final)} \\
\midrule
\texttt{instruct-no-sft-math} & - & 0.550 & 0.598 & 0.604 (150step) \\
\texttt{long-CoT-math} & 690 & 0.184 & 0.554 & 0.580 (150step) \\
\texttt{short-CoT-math} & 900 & 0.298 & 0.530 & - \\
\texttt{base-no-sft-math} & - & 0.238 & 0.582 & - \\
\texttt{RD-sft-math} & 496 & 0.544 & 0.586 & 0.604 (150 step) \\
\texttt{distill-50k-math} & 50k & 0.278 & 0.546 & 0.572 (300 step) \\
\texttt{long-CoT-temp1.2-math} & 690 & 0.184 & 0.584 & - \\
\texttt{long-CoT-hyperfit-math} & 690 & 0.426 & 0.416 & - \\
\bottomrule

\end{tabular}
\end{table*}

\newpage

\section{Proofs} \label{sec:appendix-proof}

\textbf{Assumptions in this paper:}

\begin{itemize}
    \item Our core assumption is: linearized sample effect is able to explain the effectiveness of multi-step and non-linear LLM training process.
    \item To analyze discrete gradient decent with SDE, learning rate \(\eta\) should be sufficiently small.
    \item \(N\) is sufficiently large to apply Central Limit Theorem.
    \item In initial steps of RL, by setting suitable hyperparameters, we can estimate drift term by ignoring the noise effect.

\end{itemize}

\subsection{Proof of Instantaneous Growth Rate of Accuracy} \label{subsec:app-proof-drift}

We start from the policy gradient loss:

\begin{equation} \label{equ:appendix-pg-raw}
    -\nabla_\theta L=\frac{1}{N}\sum_{a \sim \pi_\theta, s \sim D_t}[\vec \nabla_\theta \ln \pi_\theta(a,s) r(a,s)]
\end{equation}

By Central Limit Theorem, we can represent the estimation noise by Wiener Process. 

\begin{equation} \label{equ:appendix-wp}
    \mathrm{d}\theta =\mathbb{E}_{a\sim\pi_\theta, s \sim D_t}[\vec \nabla_\theta \ln \pi_\theta(a,s) r(a,s)]\mathrm{d}t+\frac{\eta}{\sqrt{N}}\mathbf{A} \mathrm{d} \vec W_t
\end{equation}

The matrix \(\mathbf{A}\) satisfies:

\begin{equation} 
    \mathbf{A}\mathbf{A}^T=\text{Cov}_{s\sim D_t, a\sim \pi_\theta}[r(a,s) \partial/\partial{\theta_i} \ln \pi_\theta(a,s), r(a,s) \partial/\partial{\theta_j} \ln \pi_\theta(a,s)]
\end{equation}

Let \(\vec A_i\) denotes the i-th column in \(\mathbf{A}\). \(W_i\) denotes the i-th component of \(\vec W_t\). We can re-write Equation \ref{equ:appendix-wp} as:

\begin{equation} \label{equ:appendix-wp-2}
    \mathrm{d}\theta =\mathbb{E}_{s\sim D_t,a\sim\pi_\theta}[\vec \nabla_\theta \ln \pi_\theta(a,s) r(a,s)]\mathrm{d}t+\frac{\eta }{\sqrt{N}}\sum_{i=1}^n\vec A_i\mathrm{d} W_{i}
\end{equation}

We define the growth rate of test reward(accuracy) in time \(t\) as \(\Psi(\theta(t))\):

\begin{equation}
    \Psi(\theta)=\mathbb{E}_{a \sim \pi_\theta, s \sim D_e}[r(a,s)]
\end{equation}

For a standard SDE \(\mathrm{d}\vec X = \vec f(\vec X,t)\mathrm{d}t + \sum_{i=1}^n\vec g_i(\vec X,t)\mathrm{d}W_i\), we can compute any \(\Psi(\vec X)\) using Ito's lemma:

\begin{align} \label{equ:appendix-sde}
    \mathrm{d}\Psi(\vec X) = & \mu(t, \vec X) \mathrm{d}t + \sigma(t, \vec X) \mathrm{d} \vec W_t \\
    = & \underbrace{[\vec \nabla_X \Psi]^\top \vec f(\vec X) \mathrm{d}t + \frac{1}{2} \sum_{i=1}^n \vec g_i(\vec X)^\top \nabla^2\Psi(\vec X) g_i(\vec X)\mathrm{d}t}_{\textbf{Drift term}} \\
    & + \underbrace{\sum_{i=1}^{n} [\vec \nabla_X \Psi]^\top \vec g_i(\vec X) \mathrm{d}W_i}_{\textbf{Noise term}}
\end{align}

We derive drift term by substituting variables in the first and second terms.

\begin{align}
    \mu(t, \Psi(\theta))= &\mathbb{E}_{s\sim D_e, a\sim\pi_\theta}[\vec \nabla_\theta \ln \pi_\theta(a,s) r(a,s)]^\top \mathbb{E}_{s\sim D_t, a\sim\pi_\theta}[\vec\nabla_\theta \ln \pi_\theta(a,s) r(a,s)]\\
     & + \frac{1}{2}\frac{\eta^2}{N}\sum_{i=1}^n\vec A_i^T\nabla^2\Psi(\vec X)\vec A_i
\end{align}

\subsection{Proof of Theorem \ref{theorem:opt-sft-policy}} \label{subsec:app-proof-t1}

The proof of Theorem \ref{theorem:opt-sft-policy} is similar to DPO \citep{rafailov_direct_2023}. Let \(\pi_\theta^\dagger(a,s)\) to be the target policy. \(\hat{\pi}_\theta^\dagger(a,s)\) to be the distribution after correctness filter. We have \(\hat{\pi}_\theta^\dagger(a,s)=\frac{\pi_\theta^\dagger(a,s)}{p_\theta^\dagger(s)}r(a,s)\). We find \(\pi_\theta^\dagger(a,s)\), which satisfies \(\mathbb{E}_{s\sim D_t, a\sim \pi_\theta^\dagger}[r(a,s)] = p_\theta^\dagger(s)\), to maximize the objective \(\mathcal{J}\). 

\begin{align}
    \mathcal{J} = & \mathbb{E}_{s\sim D_t, a\sim \hat{\pi}_\theta^\dagger}[V(a,s,\theta)r(a,s)] - \mathbb{E}_{s\sim D_t, a\sim\pi_\theta}[V(a,s,\theta) r(a,s)] \\
    & - \beta \mathbb{D}_{KL}[\pi_\theta^\dagger(a,s)||\pi_\theta(a,s)] \\
    V(a,s,\theta) = & \mathbb{E}_{s\sim D_e, a\sim\pi_\theta}[\vec \nabla_\theta \ln \pi_\theta(a,s) r(a,s)]^\top \vec \nabla_\theta \ln \pi_\theta(a,s) \\
    \hat{\pi}_\theta^\dagger(a,s)= & \frac{\pi_\theta^\dagger(a,s)}{p_\theta^\dagger(s)}r(a,s) \\
    p_\theta^\dagger(s) = & \mathbb{E}_{s\sim D_t, a\sim \pi_\theta^\dagger}[r(a,s)] \\
    \mathbb{D}_{KL}[\hat{\pi}_\theta^\dagger(a,s)||\pi_\theta(a,s)] = & \mathbb{E}_{s \sim D_t, a \sim \hat{\pi}_\theta^\dagger}[\ln \frac{\hat{\pi}_\theta^\dagger(a,s)}{\pi_\theta(a,s)}] \\
    r(a,s) = & 1 \text{ if }a\text{ is correct else } 0
\end{align}

The second term of \(\mathcal{J}\) does not depend on \(\pi_\theta^\dagger(a,s)\). We eliminate the second term and rewrite objective as:

\begin{align} \label{equ:appendix-max-j}
    \max_{\pi_\theta^\dagger} & \mathcal{J} \\
    = & \max_{\pi_\theta^\dagger} \mathbb{E}_{s\sim D_t, a\sim \hat{\pi}_\theta^\dagger}[V(a,s,\theta)r(a,s)] - \beta \mathbb{D}_{KL}[\pi_\theta^\dagger(a,s)||\pi_\theta(a,s)] \\
    = & \max_{\pi_\theta^\dagger} \mathbb{E}_{s\sim D_t, a\sim \pi_\theta^\dagger}[\frac{1}{p_\theta^\dagger(s)}V(a,s,\theta)r(a,s) - \beta \ln \frac{\pi_\theta^\dagger(a,s)}{\pi_\theta(a,s)}] \\
    = & \min_{\pi_\theta^\dagger} \mathbb{E}_{s\sim D_t, a\sim \pi_\theta^\dagger}[\ln \frac{\pi_\theta^\dagger(a,s)}{\frac{1}{Z(a,s)}\pi_\theta(a,s)\exp(\frac{1}{\beta p_\theta^\dagger(s)}  V(a,s,\theta)r(a,s))} - \ln Z(a,s)]
\end{align}

In the above we use \(r(a,s)=r^2(a,s)\). For simplicity, we denote \(\hat r(a,s, \theta)=\frac{1}{p_\theta^\dagger(s)} V(a,s,\theta)r(a,s)\). Differ from DPO, we define \(Z(a,s)\) as:

\begin{equation}  
  Z(a,s)=\left\{  
       \begin{array}{lr}  
        \frac{1}{p_\theta^\dagger(s)}\sum_{a}r(a,s)\pi_\theta(a,s)\exp(\frac{1}{\beta}\hat r(a,s,\theta)) = Z^+(s) & \text{if } r(a,s)=1  \\
        \frac{1}{1-p_\theta^\dagger(s)}\sum_{a}(1-r(a,s))\pi_\theta(a,s) = Z^-(s) & \text{if } r(a,s)=0  \\
       \end{array}
  \right.  
\end{equation}

Then we define \( \pi^*_\theta(a,s) = \frac{1}{Z(a,s)}\pi_\theta(a,s)\exp(\frac{1}{\beta} \hat r(a,s,\theta))\). We have \( \pi^*_\theta(a,s) \geq 0\), and it is a valid distribution:

\begin{align}
    \sum_{a}\pi^*_\theta(a,s) = & \sum_{a}r(a,s)\pi^*_\theta(a,s) + \sum_{a}(1-r(a,s))\pi^*_\theta(a,s) \\
    = & \frac{1}{Z^+(s)}\sum_a r(a,s) \pi_\theta(a,s)\exp(\frac{1}{\beta} \hat r(a,s,\theta)) + \frac{1}{Z^-(s)}\sum_a (1-r(a,s)) \pi_\theta(a,s) \\
    = & p_\theta^\dagger(s) + (1 - p_\theta^\dagger(s)) \\
    = & 1
\end{align}

We can also check the accuracy constraint is satisfied: \( \sum_{a}r(a,s)\pi^*_\theta(a,s) = p_\theta^\dagger(s)\).

To eliminate \(\ln Z(a,s) \), we prove that \(\mathbb{E}_{s\sim D_t, a\sim \pi_\theta^\dagger}[\ln Z(a,s)]\) does not depend on \(\pi_\theta^\dagger\): 

\begin{align}
\mathbb{E}_{s\sim D_t, a\sim \pi_\theta^\dagger}[\ln Z(a,s)] = & \sum_{s,a}\pi_\theta^\dagger(a,s) \ln Z(a,s) \\
    = & \sum_{s,a}\pi_\theta^\dagger(a,s)r(a,s)\ln Z^+(s) + \sum_{s,a}\pi_\theta^\dagger(a,s)(1-r(a,s))\ln Z^-(s) \\
    = & \sum_s [\ln Z^+(s) \sum_a \pi_\theta^\dagger(a,s)r(a,s) + \ln Z^-(s)\sum_a \pi_\theta^\dagger(a,s)(1-r(a,s))] \\
    = & \sum_s [p_\theta^\dagger(s)\ln Z^+(s) + (1-p_\theta^\dagger(s)) \ln Z^-(s)]
\end{align}

Therefore, we can simplify the last line in Equation \ref{equ:appendix-max-j} and finally get:

\begin{align}
    \max_{\pi_\theta^\dagger} \mathcal{J} = & \min_{\pi_\theta^\dagger} \mathbb{E}_{s\sim D_t, a\sim \pi_\theta^\dagger}[\ln \frac{\pi_\theta^\dagger(a,s)}{\pi_\theta^*(a,s)}] \\
    = & \min_{\pi_\theta^\dagger} \mathbb{D}_{KL}[\pi_\theta^\dagger(a,s) || \pi_\theta^*(a,s)]
\end{align}

The KL distance achieves global minimum 0 if and only if \(\pi_\theta^\dagger(a,s)=\pi_\theta^*(a,s)\). We prove that \(\pi_\theta^*(a,s)\) is the optimal policy.

For any \(a_1, a_2\) and \(s\in \{D_t\}\) satisfies \(r(a_1,s)=r(a_2,s)=1\), we have: 

\begin{align}
\ln{\frac{\pi_\theta^*(a_1,s)}{\pi_\theta^*(a_2,s)}} = & \ln \frac{\frac{1}{Z^+(s)}\pi_\theta(a_1,s)\exp(\frac{1}{\beta}\hat r(a_1,s,\theta))}{\frac{1}{Z^+(s)}\pi_\theta(a_2,s)\exp(\frac{1}{\beta}\hat r(a_2,s,\theta))}\\
= &\frac{1}{\beta p_\theta^\dagger(s)}(V(a_1,s,\theta)-V(a_2,s,\theta)) +  \ln\frac{\pi_\theta(a_1,s)}{\pi_\theta(a_2,s)}
\end{align}

To prove Corollary 1, note that for any new \(p_\theta^\dagger(s) > 0\), we can always find another \(\beta'\) and \(p_\theta'(s)\) to keep \(\frac{1}{\beta} \hat r(a,s,\theta) = \frac{1}{\beta p_\theta^*(s)} V(a,s,\theta)r(a,s)\) unchanged. The filtered distribution \( \hat{\pi}_\theta^\dagger(a,s) = r(a,s) \frac{\pi_\theta(a,s)\exp(\frac{1}{\beta} \hat r(a,s,\theta))}{ p_\theta^*(s)Z^+(s)}\) contains no single \(p_\theta^*(s)\). Therefore, the optimal distribution keep unchanged.

\subsection{Proof of Theorem \ref{theorem:rl-effect}} \label{subsec:app-proof-t2}

Given a policy \(\pi_\theta^*(a,s)\) in RL, the distribution after filtering is \(\hat \pi_\theta^*(a,s)\). We use filtered data to fine-tune initial policy \(\pi_\theta(a,s)\). We define dataset effect as:

\begin{align}
V(\theta, \theta^*, D_t) = &\mathbb{E}_{s\sim D_t, a\sim \hat \pi_\theta^*}[V(a,s,\theta)r(a,s)] \\
= & \mathbb{E}_{s\sim D_t, a\sim \pi_\theta^*}[\frac{1}{p_\theta^*(s)}V(a,s,\theta)r(a,s)]
\end{align}

Note that sample effect \(V(a,s,\theta)\) and dataset effect \(V(\theta, \theta^*, D_t)\) have different input variables. The growth rate of training set dataset effect can be expressed as: 

\begin{align} \label{equ:appendix-dataset-effect-1}
    \hat{\mu}(t, V(\theta, \theta^*, D_t)) = & [\nabla_{\theta^*}V(\theta, \theta^*, D_t)]^\top\mathbb{E}_{s\sim D_t, a\sim \pi_\theta^*}[\nabla_{\theta^*}\ln \pi_\theta^*(a,s)r(a,s)] \\
    = & \mathbb{E}_{s\sim D_t, a\sim \pi_\theta^*}[\frac{1}{p_\theta^*(s)}\nabla_{\theta^*}\ln \pi_\theta^*(a,s) V(a,s,\theta)r(a,s)]^\top \mathbb{E}_{s\sim D_t, a\sim \pi_\theta^*}[\nabla_{\theta^*}\ln \pi_\theta^*(a,s)r(a,s)] \\
    = & \mathbb{E}_{s\sim D_t, a\sim \pi_\theta^*}[\frac{1}{p_\theta^*(s)}V(a,s,\theta^*)V(a,s,\theta)r(a,s)]
\end{align}

Set \(\theta=\theta^*\) to get the instantaneous growth rate of dataset effect. Continue to simplify Equation \ref{equ:appendix-dataset-effect-1}:

\begin{align} \label{equ:appendix-dataset-effect-2}
\hat{\mu}(t, V(\theta, \theta^*, D_t)) = & \mathbb{E}_{s\sim D_t, a\sim \pi_\theta}[\frac{1}{p_\theta(s)}V^2(a,s,\theta)r(a,s)]
\end{align}

Since \(\frac{1}{p_\theta(s)} \geq 1\) and \(r(a,s)=r^2(a,s)\), we have:

\begin{align}
\hat{\mu}(t, V(\theta, \theta^*, D_t)) \geq & \mathbb{E}_{s\sim D_t, a\sim \pi_\theta}[V^2(a,s,\theta)r^2(a,s)]
\end{align}

By \(E[X^2] \geq E^2[X]\), we have: 

\begin{align}
\hat{\mu}(t, V(\theta, \theta^*, D_t)) \geq & \mathbb{E}_{s\sim D_t, a\sim \pi_\theta}[V^2(a,s,\theta)r^2(a,s)] \\
\geq & ||\mathbb{E}_{s\sim D_t, a\sim \pi_\theta}[V(a,s,\theta)r(a,s)]||^2 \\
= & \hat \mu^2(t, \Psi(\theta))
\end{align}

\newpage
\section{Training Hyperparameters and Implementations} \label{sec:training-details}

Our RL implementation is based on VeRL \citep{sheng_hybridflow_2024} and vLLM \citep{kwon_efficient_2023}. We use 2xA800 80G for all experiments.

\subsection{Supervised Fine-tuning}

In SFT stage, we use Adam optimizer without weight decay(\(\beta_1 = 0.9, \beta_2 = 0.999\)). On K\&K dataset, we use batch size 64, learning rate 1e-5, cosine scheduler and 5000 token as max length. We train every model for 2 epochs except \texttt{short-CoT-kk} for 1 epoch because its loss reached near zero within one epoch. For small-scale SFT on MATH and REASONING GYM, we use a batch size of 32 and keep other settings unchanged. 
Re-distillation has no difference with SFT in hyper-parameters except \texttt{RD-rl-kk}. Since the aim of \texttt{RD-rl-kk} is to exploit information from 1K samples, we use batch size equals to 32 to boost performance. 

For Continued PreTraining in \texttt{distill-50k-math}, we use batch size 64, 12K tokens as max length, learning rate 1e-5, cosine scheduler and 1 training epoch. We found a larger batch size and lower learning rate make CPT more stable. 

\subsection{Reinforcement Learning}

\textbf{Hyperparameters}: We list GRPO configuration as follows.

Rollout: 8 responses per question (temperature=0.7, max length=4096), 128 random selected problems per rollout. Exceptionally, we use max length=8192 for \texttt{re-distillation-rl-kk} and 256 problems per rollout for all experiments on REASONING GYM. In Llama-3.2-3B RL experiments, we use max response length=2048 to reduce VRAM usage.

Policy update: Single batch(1024-sample for MATH and KK, 2048-sample for REASONING GYM), clip ratio=0.2, gradient norm=1.0

Optimizer: Adam (\(\beta_1=0.5\), \(\beta_2=0.999\))

Learning rates: 5.0e-6 (K\&K and REASONING GYM) or 2.5e-6 (MATH), with 4 warm-up steps

Other settings: Sequence packing in computing logprobs and updating policy.

\begin{figure}[h]
\centering
\includegraphics[width=0.4\linewidth]{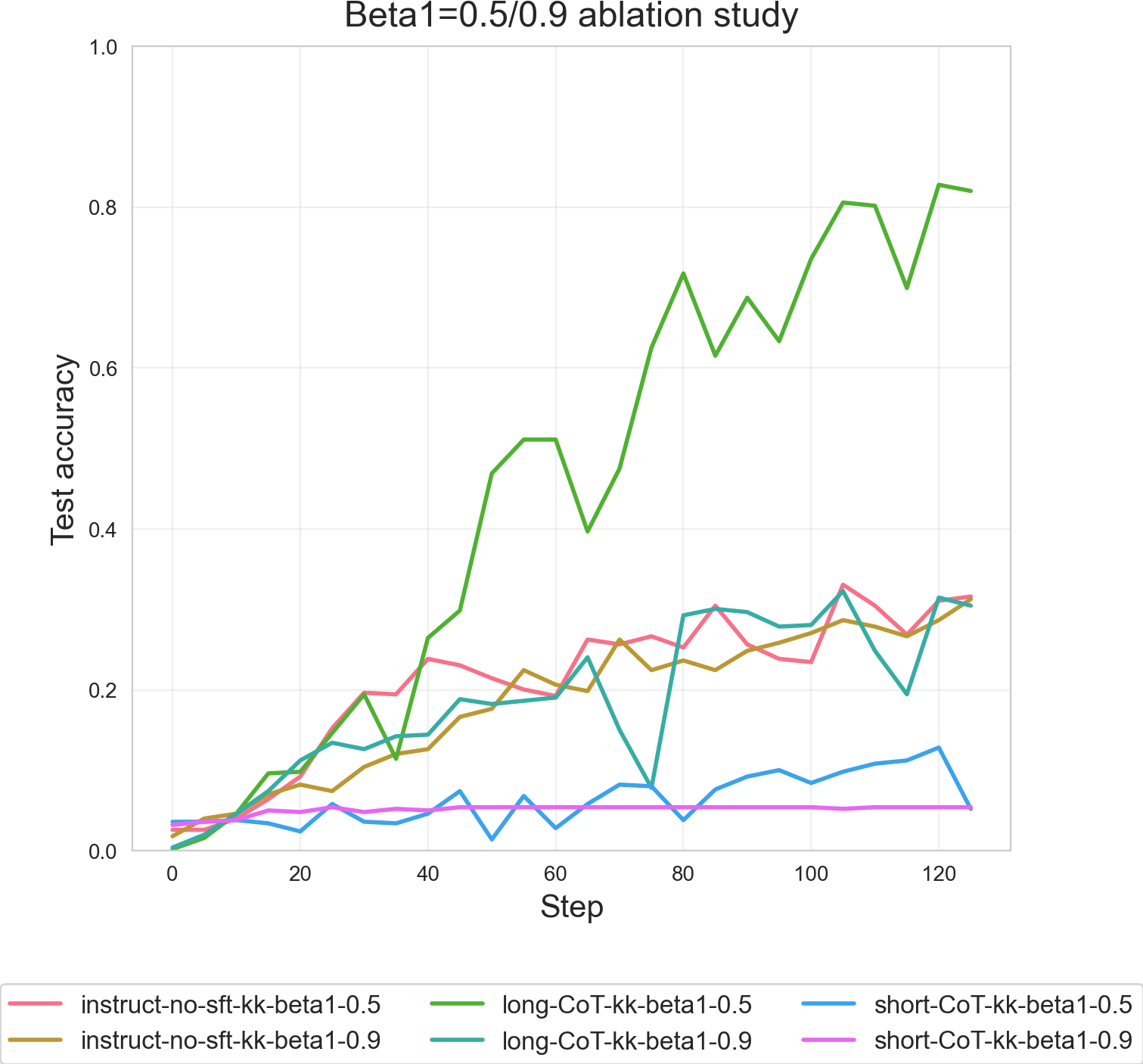}
\caption{Test accuracy for hyper-parameter ablation study: \(\beta_1=0.9\) vs \(\beta_1=0.5\). We observed that \(\beta_1=0.5\) shows equal or better performance than \(\beta_1=0.9\) under three settings.}
\label{fig:hyper-beta1}
\end{figure}

\textbf{Training Algorithm}: We adapt GRPO and eliminate the KL divergence constraint, following DAPO's approach \citep{yu_dapo_2025}. To improve RL stability, we set the training batch size equal to the rollout buffer size, ensuring sufficient valid samples per update. Empirical results demonstrate better performance with lower \(\beta_1\) in Adam optimizer parameters (\(\beta_1=0.5, \beta_2=0.999\)), particularly when encountering lengthy outputs that frequently trigger clipping. We conduct ablation experiments for \(\beta_1\) in Adam optimizer on \texttt{long-CoT-kk}, \texttt{short-CoT-kk} and \texttt{instruct-no-sft-kk}. In Figure \ref{fig:hyper-beta1}, all three experiments show consistent improvement by tuning \(\beta_1\) from 0.9 to 0.5. Therefore, we use this setting in the following experiments.

\textbf{Reward Function}: On K\&K and MATH, we use binary reward function which only outputs 0 or 1. This reward function requires the end-of-sentence token to give correctness reward. Responses exceeding output length will be assigned to 0 reward despite its correctness. We observed that 1.5B models tend to endlessly repeat final answer and fill the rest token lengths, which prohibits natural response length growth in some cases. This design prevents excessive repetition by requiring an end-of-sentence token. 
The K\&K dataset employs strict matching between extracted name-identity lists and ground truth. The reward function for checking MATH answers is implemented based on \citet{zeng_simplerl-zoo_2025} and \citet{yang_qwen2_2024}. We slightly modify it to check the answer in final boxed\(\{\}\). Both MATH and K\&K reward functions assign zero reward to responses exceeding length limits or containing no boxed answers. We exclude format reward to prevent instability and keep simplicity. For REASONING GYM reward, we wrap the provided reward functions and directly return zero if there is no stop token or no boxed answer.

\newpage
\section{Evaluation Metrics}  \label{sec:appendix-metrics}

We evaluate model performance using \emph{pass@1} accuracy on held-out test sets. For R1-distilled models, we employ the recommended configuration from DeepSeek: temperature=0.6, top\_p=0.95, and system prompt removal. Other evaluations use greedy decoding. Following \citet{zeng_simplerl-zoo_2025} and \citet{chern_generative_2023}, when evaluating base models, we employ question-answer prompts to mitigate poor instruction-following capability. All reported results are zero-shot, which may yield minor discrepancies compared to Qwen technical report \citep{yang_qwen2_2024} due to metric differences. 

In Table \ref{tab:kk-acc}, we set max response token limit to 8K for DeepSeek-V3-0324 and Deepseek-R1-Qwen-Distill-14B/7B. 16K response tokens for Deepseek-R1 and Deepseek-R1-Qwen-Distill-32B. 4K response tokens for Qwen-2.5-1.5B Instruct models. For \text{RD-rl-kk}, we use 8K response tokens to compare its performance with the same length as DeepSeek-V3-0324. For all Llama-3.2-3B based models, including \texttt{RD-llama-gym}, \texttt{rl-llama-gym} and \texttt{RD-transfer-gym}. For other trained models, we set max response tokens as 4096.

\newpage
\section{Experimental Verifications} \label{sec:exp-verify}

\begin{figure}[h]
  \centering
  \subfigure{\includegraphics[width=0.35\textwidth]{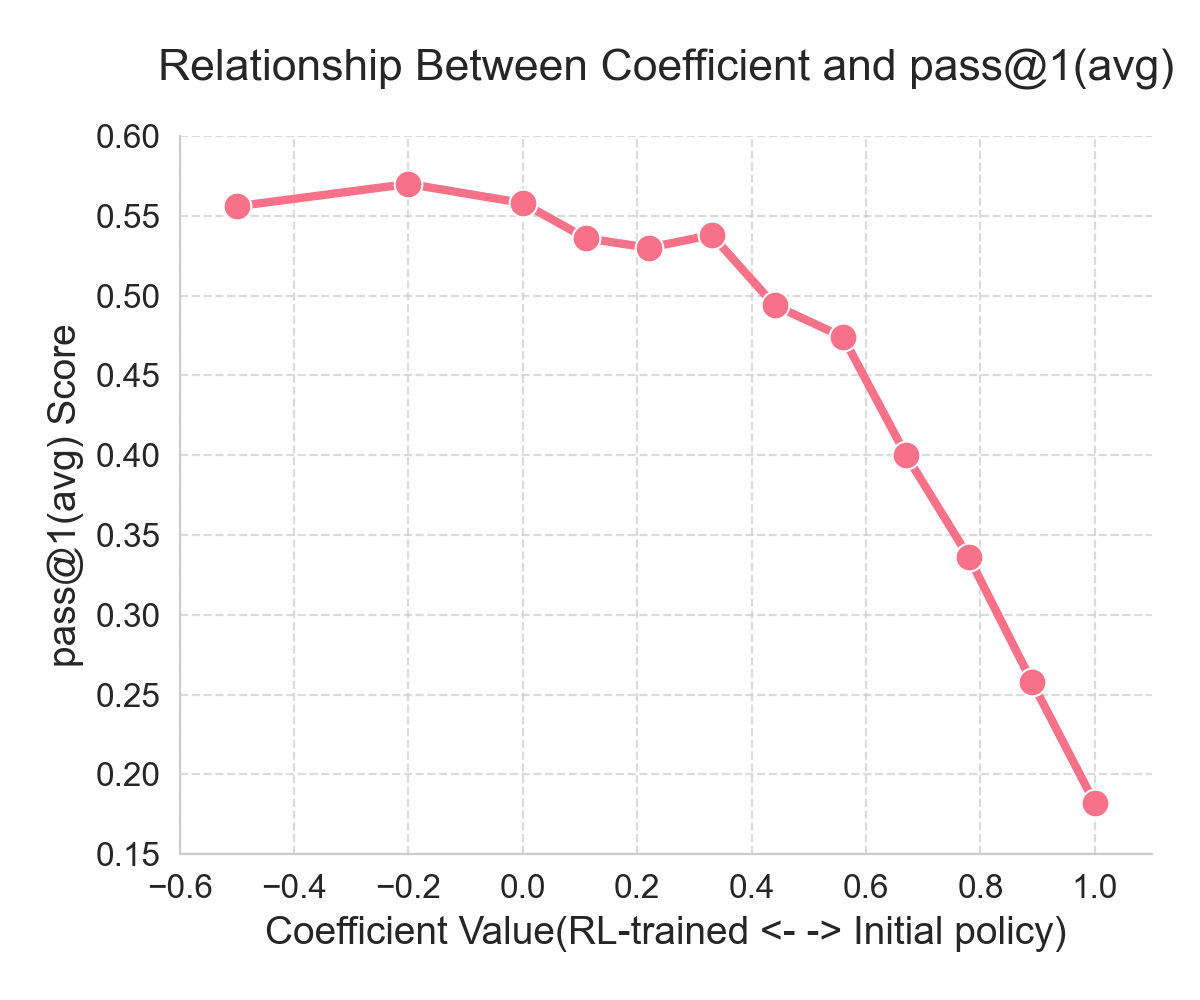}}
  \subfigure{\includegraphics[width=0.3\textwidth]{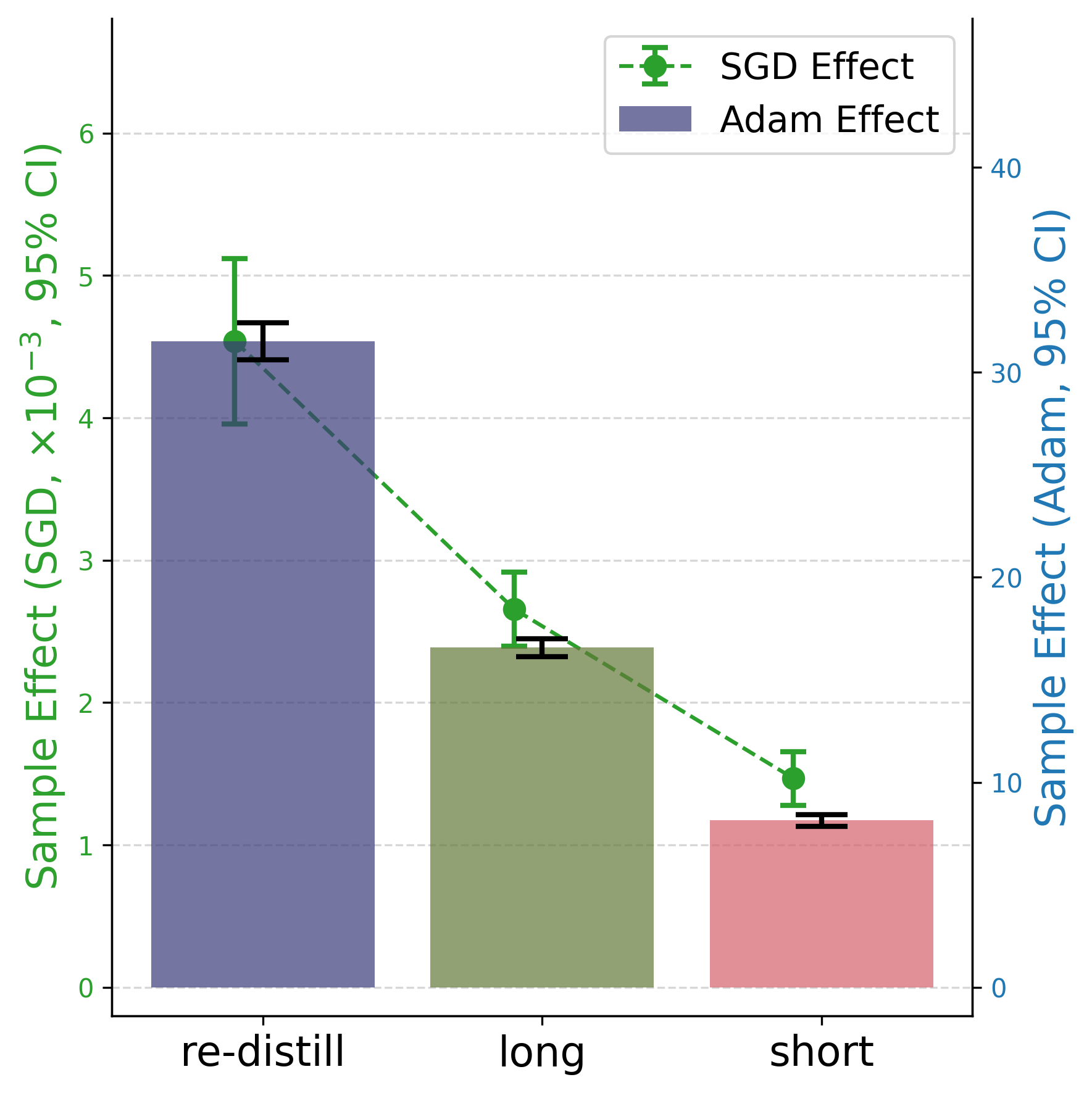}}
  \subfigure{\includegraphics[width=0.3\textwidth]{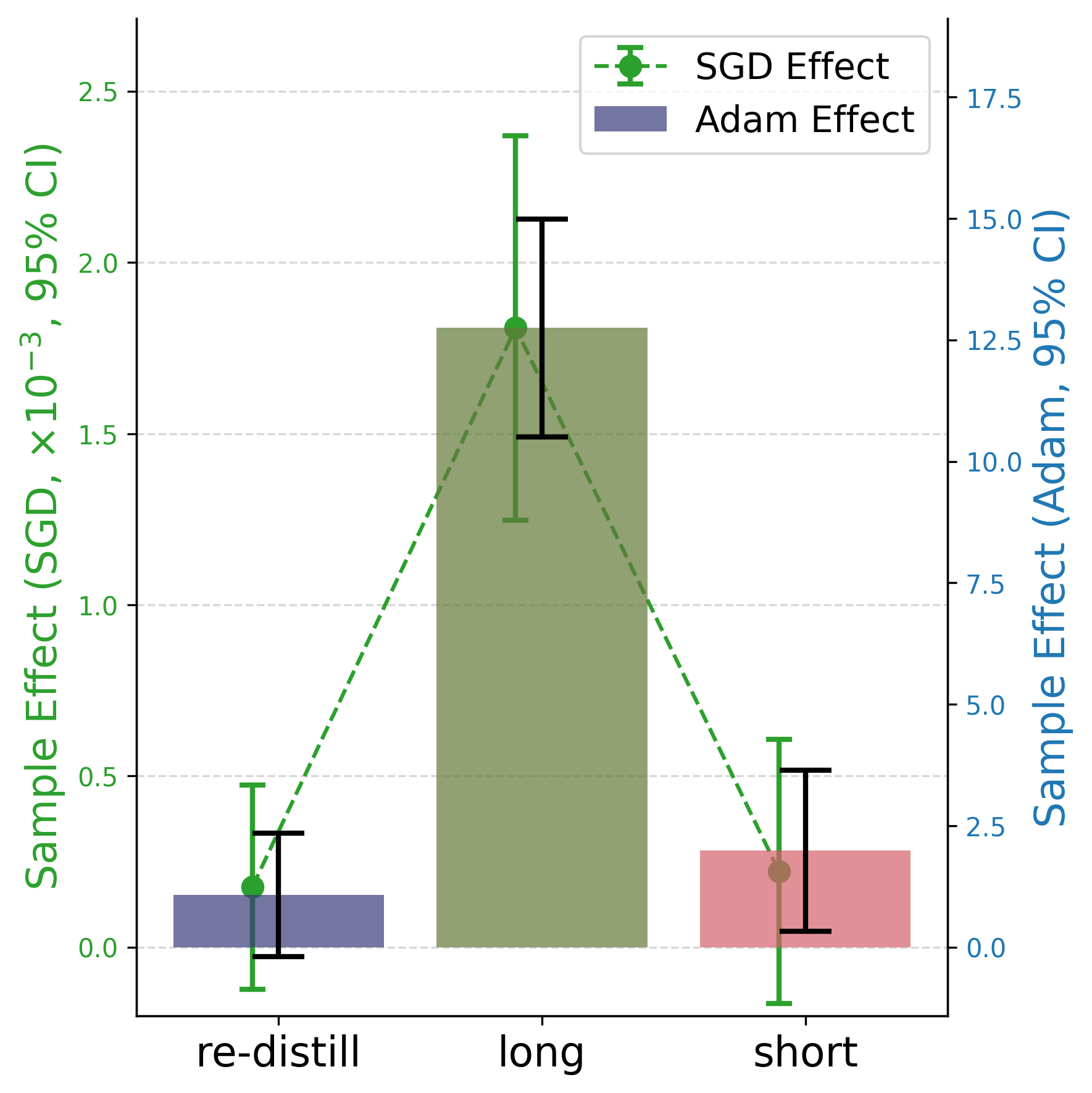}}
  \caption{\textbf{Left}: Test accuracy smoothly changes when interpolating from \texttt{long-CoT-math} initial policy to the 50-step checkpoint. \textbf{Middle}: Average sample effect(dashed green line: SGD effect, bar: Adam effect, error bar: 95\% CI) of different SFT dataset measured on Qwen-2.5-1.5B base. For each SFT model, we randomly select 500 samples from training set. \textbf{Right}: Average sample effect of rollout samples measured on RL initial policies(SFTed model). For each model, we randomly select 2000 samples from replay buffer within 4 steps.}
  \label{fig:verify}
\end{figure}

We perform further investigation to verify if the effectiveness of re-distillation can be explained by boosted sample effect. 

\textbf{RL has smooth loss landscapes}. A key requirement for successful re-distillation is a smooth RL optimization landscape. We analyze this through parameter space interpolation between initial (\(\pi(\theta_{old})\)) and RL-optimized (\(\pi(\theta)\)) policies, constructing intermediate models \(\pi(\theta_\lambda) = \pi(\lambda \theta_{old} + (1-\lambda) \theta)\) for \(\lambda \in [0,1]\). Figure \ref{fig:verify} (Left) demonstrates a smooth accuracy transition, which suggests the feasibility of approximating RL improvements without precisely replicating its gradient trajectory. 

\textbf{Estimated sample effect matches training performance}. According to Equation \ref{equ:mu-approx-rl} and Equation \ref{equ:mu-approx-sft}, better sample effect should boost training efficiency. We verify it by computing sample effects from definition. Specifically, we use normed parameter delta \(\Delta \theta = \theta_{25} - \theta_0\) between Qwen-2.5-1.5B base \(\theta_{0}\) and \texttt{base-no-sft-math} 25-step checkpoints \(\theta_{25}\) to estimate the policy gradient \(\mathbb{E}_{s\sim D_e, a\sim\pi_\theta}[\vec \nabla_\theta \ln \pi_\theta(a,s) r(a,s)]\). Although it is straight forward to compute policy gradient by its definition, we need an unbiased reference to compare dataset effect for different experiments. The parameter delta is also effective enough because \texttt{base-no-sft-math} shows rapid accuracy growth within 25 steps, which guarantees valid information in \(\Delta \theta\). 

Specifically, we compute sample effect under both SGD and Adam settings. The SGD sample effect is defined in Definition \ref{def:sample-effect}. Adam sample effect replaces the last gradient term according to Adam optimizer \citep{kingma_adam_2014}. Refer to Section \ref{sec:appendix-verify} for more details. In Figure \ref{fig:verify} (Middle), we show the estimated SFT effectiveness on different datasets. Re-distilled data in \texttt{RD-sft-math} shows 2x efficiency than long CoT data(\texttt{long-CoT-math}). This explains why re-distillation has superior effect. However, this method failed to predict the correct effect between long CoT and short CoT, it is probably because models may have long response length after distilled from DeepSeek-R1 and tend to exceed max length limit in evaluation.

We also compute the sample effect for each initial policy in RL experiments. We compute dataset effect based on initial policy(i.e. SFTed model) to verify the growth rate of test accuracy in initial steps of RL. In Figure \ref{fig:verify} (Right), \texttt{long-CoT-math} shows the highest RL efficiency. Notably, \texttt{RD-sft-math} has low sample effect because the model is almost converged at the beginning. 

\newpage
\section{More Investigations about Reasoning Patterns}

\subsection{Can LLM Progressively Improve Through Stable Reasoning Patterns?} \label{subsec:more-sft}

We investigate whether small LLMs can achieve progressive self-improvement and maintain stable reasoning patterns with more SFT. We take two approaches: extended SFT and HyperFitting. Test performances are listed in Table \ref{tab:math-acc}. Refer to Figure \ref{fig:appendix-math} for training statistics.

Firstly, we scale up supervised fine-tuning with 50K high-quality reasoning traces from OpenR1-Math-220K, which are distilled from DeepSeek-R1 on NuminaMath 1.5 \citep{face_open_2025,li_numinamath_2024}. We keep solutions no longer than 16K tokens and randomly pick 50K samples to obtain SFTed model(\texttt{distill-50k-math}). Then, the SFTed model receives small-scale SFT and RL stage as same as \texttt{long-CoT-math}. As a result, RL-trained \texttt{distill-50k-math} has worse learning curve than \texttt{long-CoT-math}. Performance plateaus with reward improvements requiring twice as many steps as baseline models to achieve comparable accuracy.

Second, we perform HyperFitting \citep{carlsson_hyperfitting_2024,ye_limo_2025} on \texttt{long-CoT-math} through 15-epoch training to train \texttt{long-CoT-hyperfit-math}, achieving near-zero SFT loss. Although the SFTed model shows reduced repetition after SFT, it exhibits severely constrained exploration and inefficient reward curve during RL stage.

\subsection{Why SFT Affects Long-Term Exploration} \label{subsec:exploration}

Our hypothetical analysis only explains how SFT influence RL at initial steps. It remains a question that why RL can not get rid of initial output modes by exploration. We propose a new hypothesis to explain this: During auto-regression, early tokens face ambiguous credit assignment until later tokens approach optimum. For instance, deep reasoning behavior gain little competitive advantage unless LLM learned to generate stop tokens, because  in our experiments reward is always zero without a correct stop token. Hence, initial tokens may converge slower than subsequent ones.

\begin{figure}[h]
  \centering
  \subfigure{
  \includegraphics[width=0.4\textwidth]{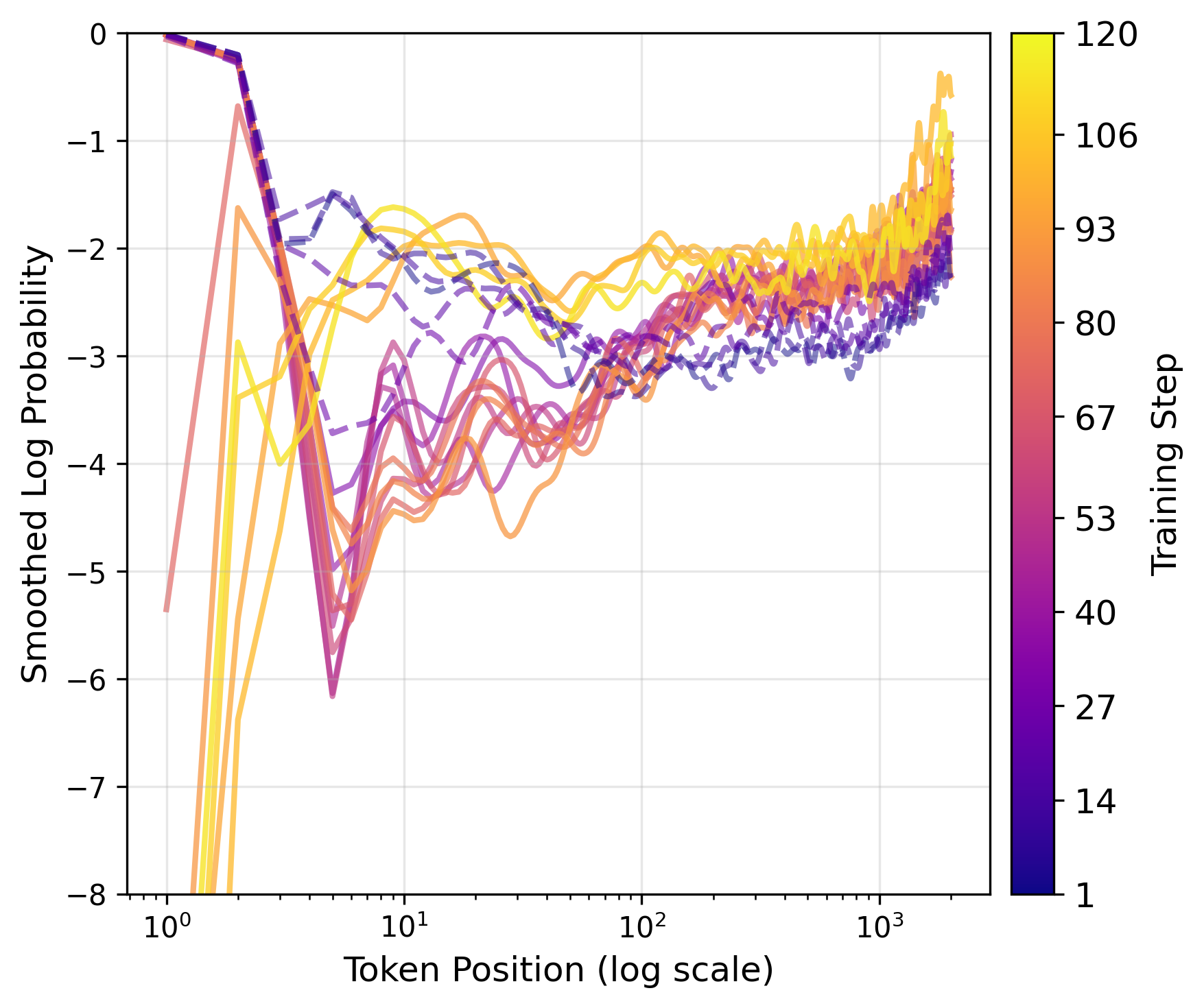}}
  \subfigure{
  \includegraphics[width=0.4\textwidth]{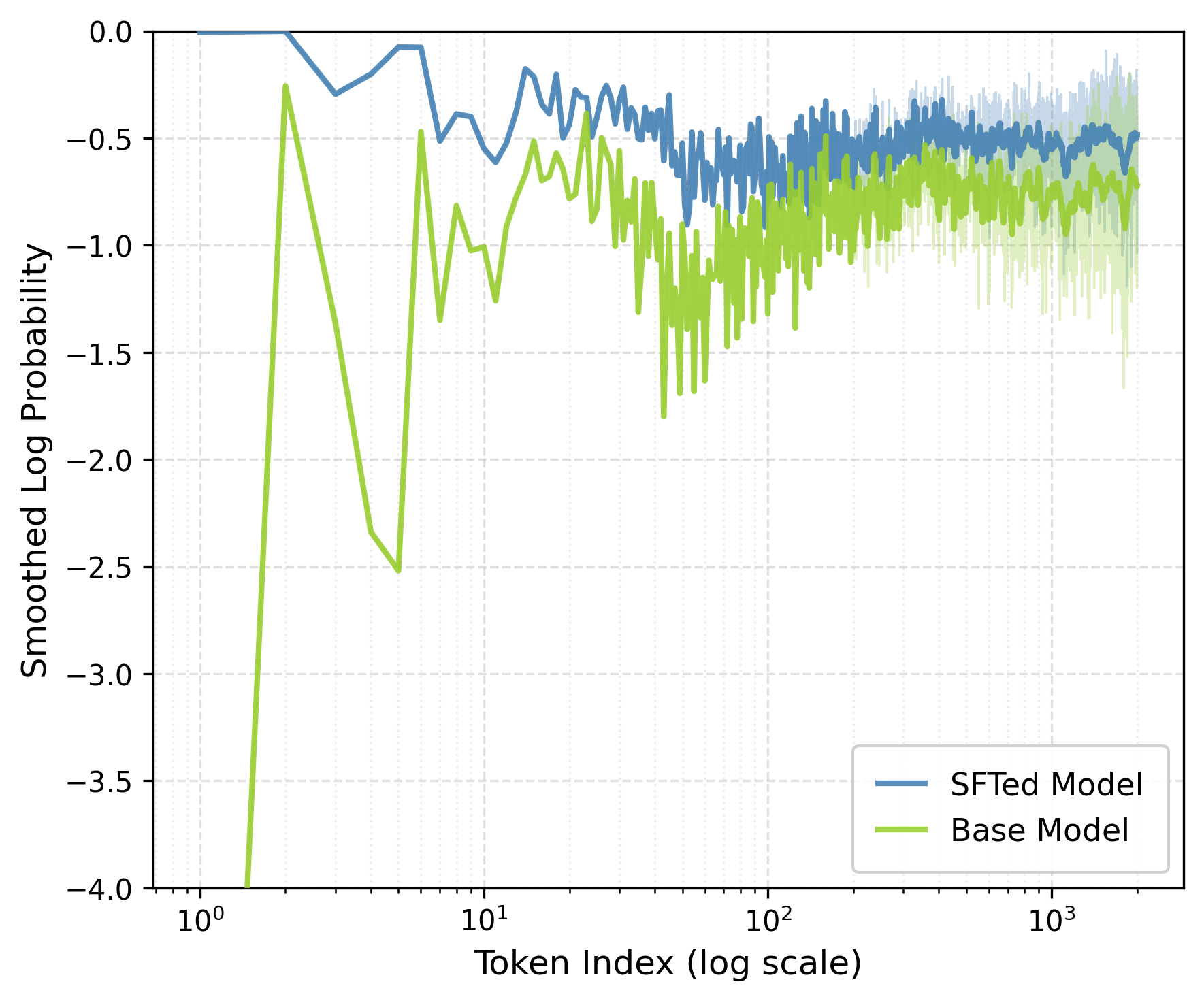}}
  \caption{\textbf{Divergent Exploration Patterns in RL and SFT}. \textbf{Left:} Position-wise logprobs from \texttt{long-CoT-math} RL rollouts evaluated by its initial policy. \textbf{Right:} Logprob distributions before/after SFT on identical data, evaluated by SFTed model of \texttt{long-CoT-math}.}
  \label{fig:logprob}
\end{figure}

To verify this hypothesis, we record position-wise logprobs in response tokens. Using initial policy, we evaluate token probabilities across different RL training steps. Then we compute the 1\% lowest logprobs per position as indicators for backward information propagation. Aggregating these measurements reveals a clear pattern in Figure \ref{fig:logprob}(Left): as RL progresses, the token position of the lowest probability point moves from terminal to initial tokens. Notably, terminal tokens exhibit rising probabilities in the converged policy, demonstrating \textbf{RL's tendency to redistribute uncertainty from final to initial tokens}.

For contrast, we examine pre-/post-SFT logprob distributions across response positions. Figure \ref{fig:logprob}(Right) reveals that SFT induces substantially larger shifts in early tokens. This effect only appear in RL when it is near convergence. Later tokens show similar trends in both pre-/post-SFT models, suggesting their distributions are shaped through in-context learning. \textbf{This demonstrates RL's inherent difficulty in modifying initial token distributions compared to SFT}, highlighting SFT's importance for effective exploration.

\begin{figure}[h]
    \centering
    \subfigure{\includegraphics[width=0.3\textwidth]{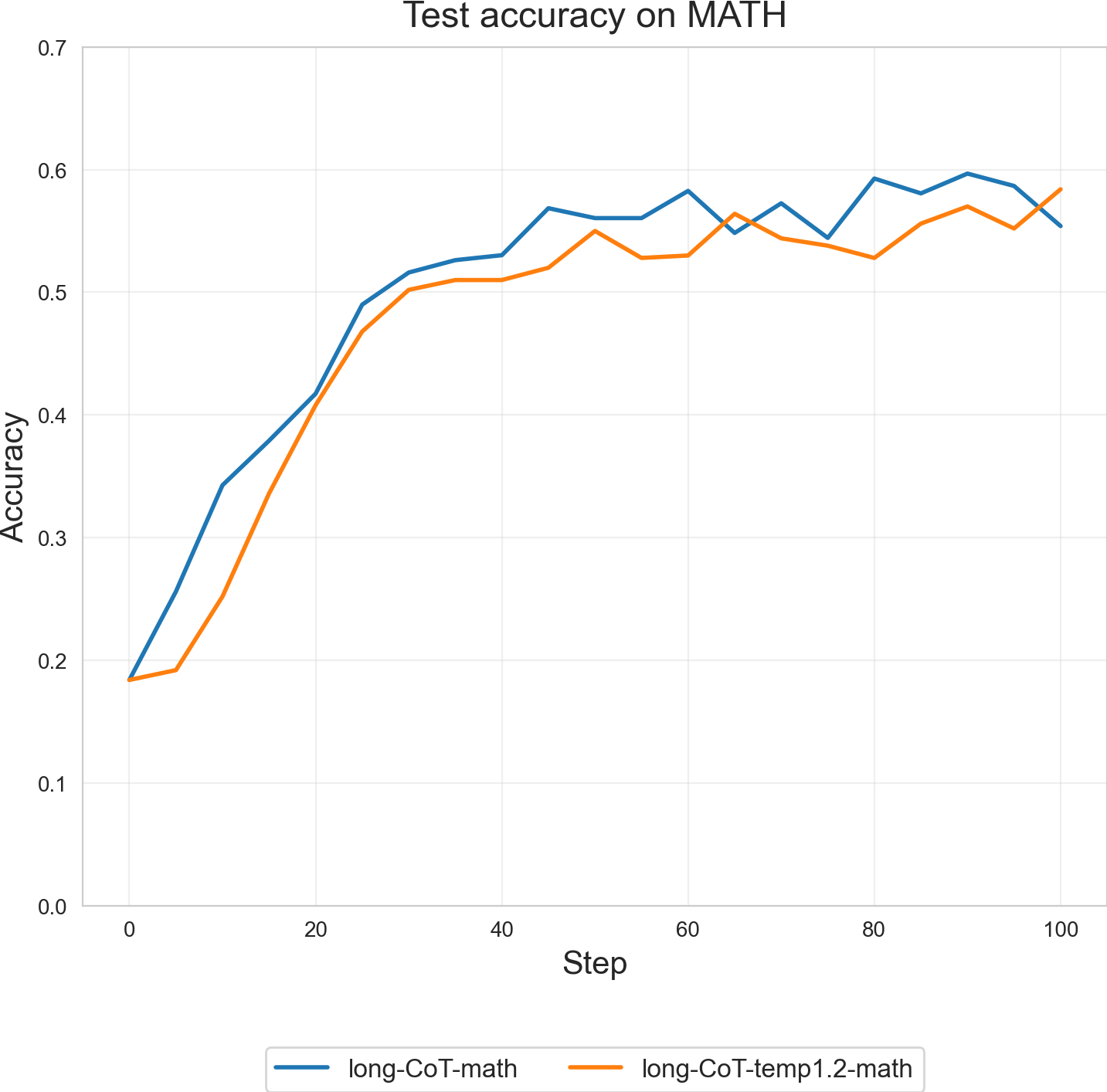}}
    \subfigure{\includegraphics[width=0.3\textwidth]{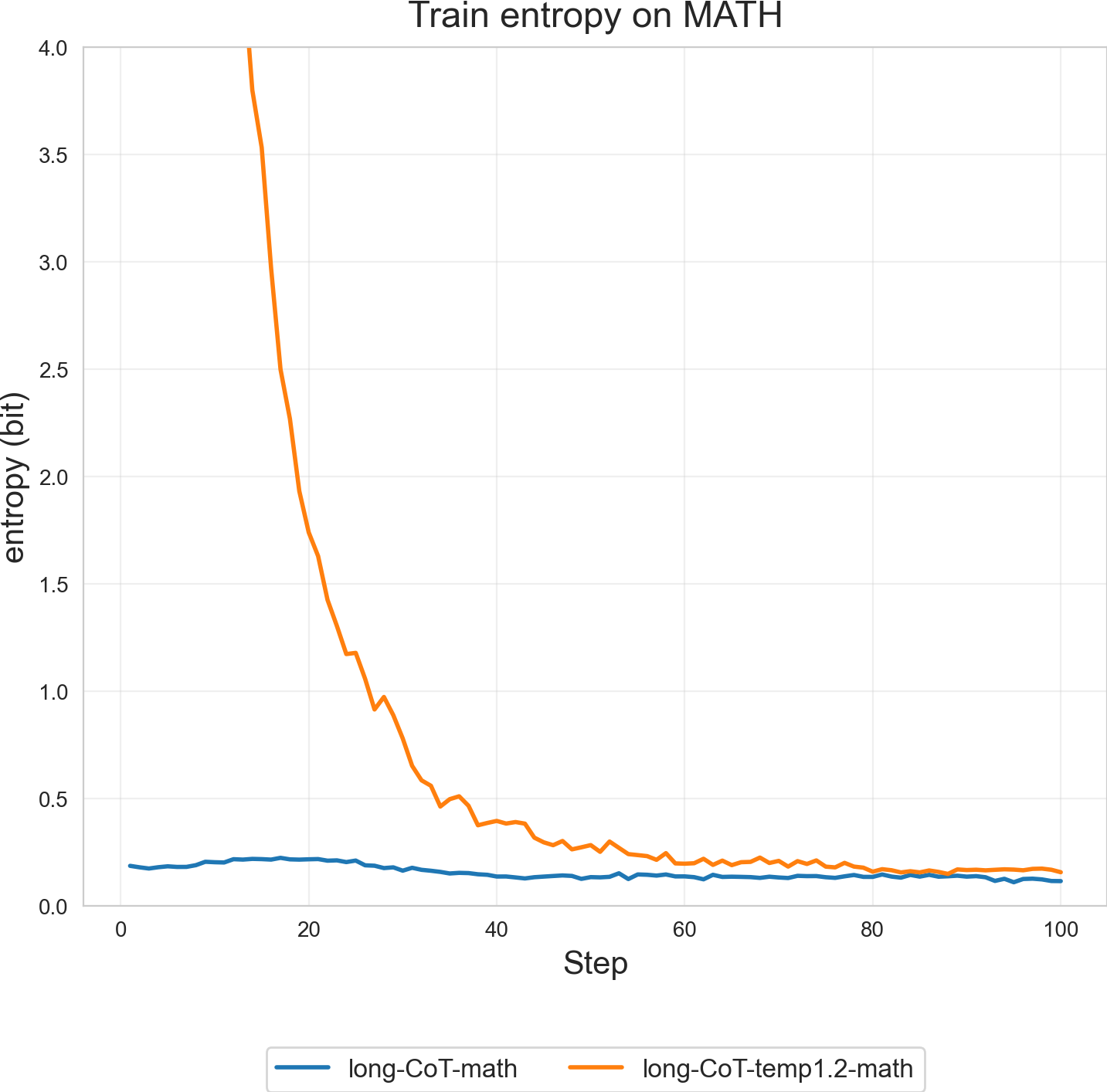}}
    \caption{\textbf{Left}: Increasing temperature will not bring substantial improvement. Two models share the same GRPO recipe except different rollout temperatures. \textbf{Right}: Increasing temperature leads to significantly higher entropy in initial steps. }
    \label{fig:appendix-temp-math}
\end{figure}

We further investigate the effect of randomness. We re-run \texttt{long-CoT-math} with an elevated RL temperature(1.2 instead of 0.7) and received similar learning trajectories, suggesting simply increase randomness cannot replace SFT's exploration benefits. As shown in Figure \ref{fig:appendix-temp-math}, the test accuracies of \texttt{long-CoT-math} and \texttt{long-CoT-temp1.2-math} have similar trends. However, raising temperature has negative effects, as it significantly increases initial exploration entropy.

\newpage
\section{Details in Empirical Verification} \label{sec:appendix-verify}

We use delta of parameters \(\Delta\theta\) as policy gradient approximation. We compute SFT effect with SGD \(V_{\text{SGD}}\) as described in Equation \ref{equ:sgd-effect}. \(|a|\) is the response token length.

\begin{align} \label{equ:sgd-effect}
\vec g_t = & \frac{1}{N} \sum_{i=1}^{N}\vec \nabla_\theta \ln \pi_\theta(a,s) / |a| \\
V_{\text{SGD}} = & - \frac{\vec g_t}{||\vec g_t||}^\top \vec \Delta\theta
\end{align}

We compute SFT effect with Adam (\(V_{\text{Adam}}\)) by Equation \ref{equ:adam-effect}.

\begin{align} \label{equ:adam-effect}
    &\vec{m}_t = \beta_1 \vec{m}_{t-1} + (1 - \beta_1) \frac{\vec g_t}{||\vec g_t||} \\
    &\vec{v}_t = \beta_2 \vec{v}_{t-1} + (1 - \beta_2) \frac{\vec g_t}{||\vec g_t||} \odot \frac{\vec g_t}{||\vec g_t||}  \\
    &\hat{\vec{m}}_t = \frac{\vec{m}_t}{1 - \beta_1^t} \\
    &\hat{\vec{v}}_t = \frac{\vec{v}_t}{1 - \beta_2^t} \\
    &V_{\text{Adam}} = -\left[ \frac{\hat{\vec{m}}_t}{\sqrt{\hat{\vec{v}}_t} + \epsilon} \right]^\top \vec{\Delta\theta}
\end{align}

We use the following equation to compute sample effect in RL. \(A(a,s)\) is the advantage function computed in RL stage. Replace \(\vec g_t\) in Equation \ref{equ:sgd-effect} and Equation \ref{equ:adam-effect} to compute sample effect with SGD or Adam.

\begin{align} \label{equ:rl-effect}
\vec g_t = & \frac{1}{N} \sum_{i=1}^{N}\nabla_\theta \ln \pi_\theta(a,s) A(a,s) / |a| \\
\end{align}

\newpage
\section{Detailed Reinforcement Learning Statistics} \label{sec:appendix-stats}

\subsection{Detailed Statistics on K\&K dataset} \label{subsec:appendix-stats-kk}

\begin{figure}[h]
  \centering
  \subfigure{\includegraphics[width=0.3\textwidth]{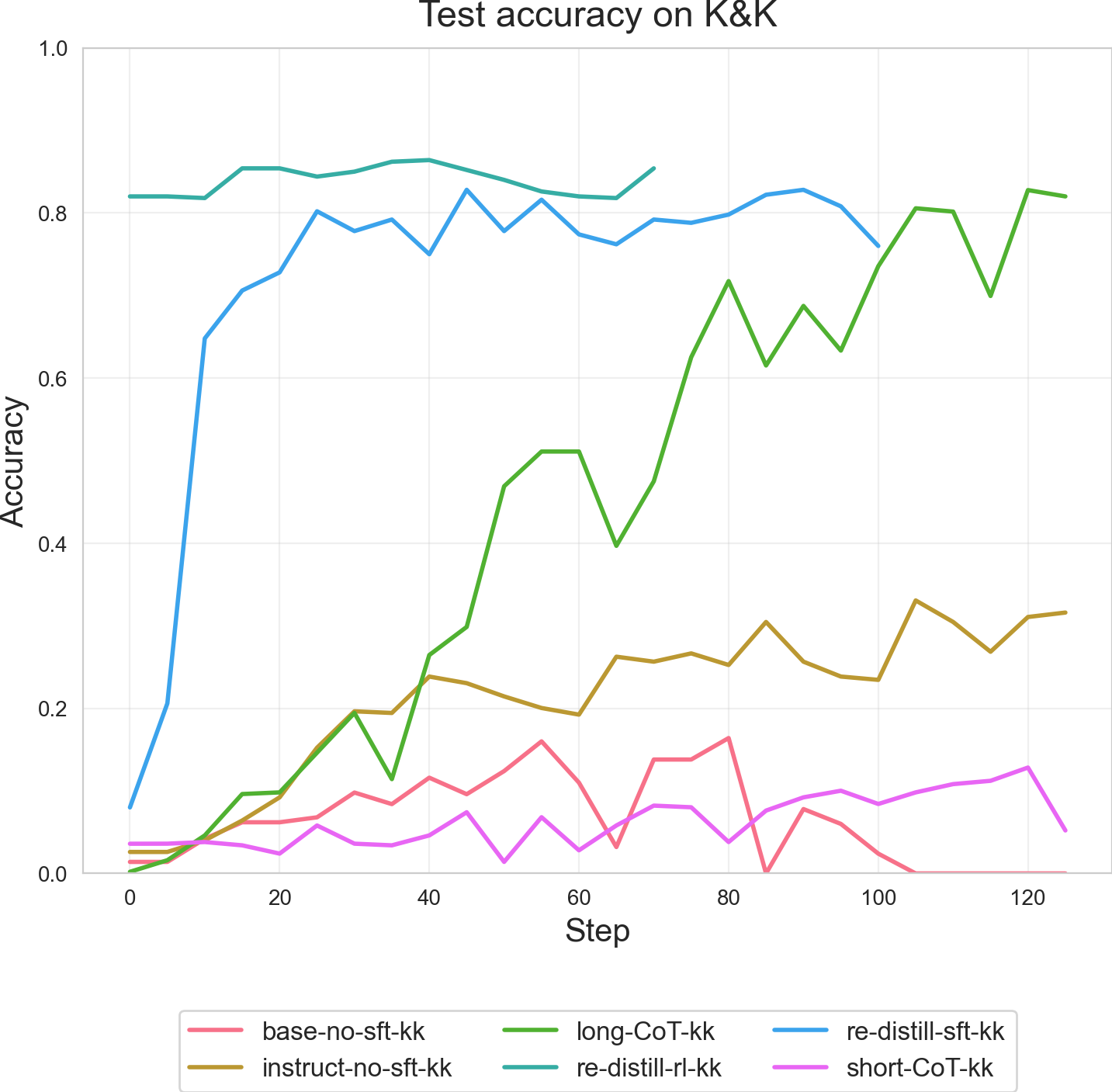}}
  \subfigure{\includegraphics[width=0.3\textwidth]{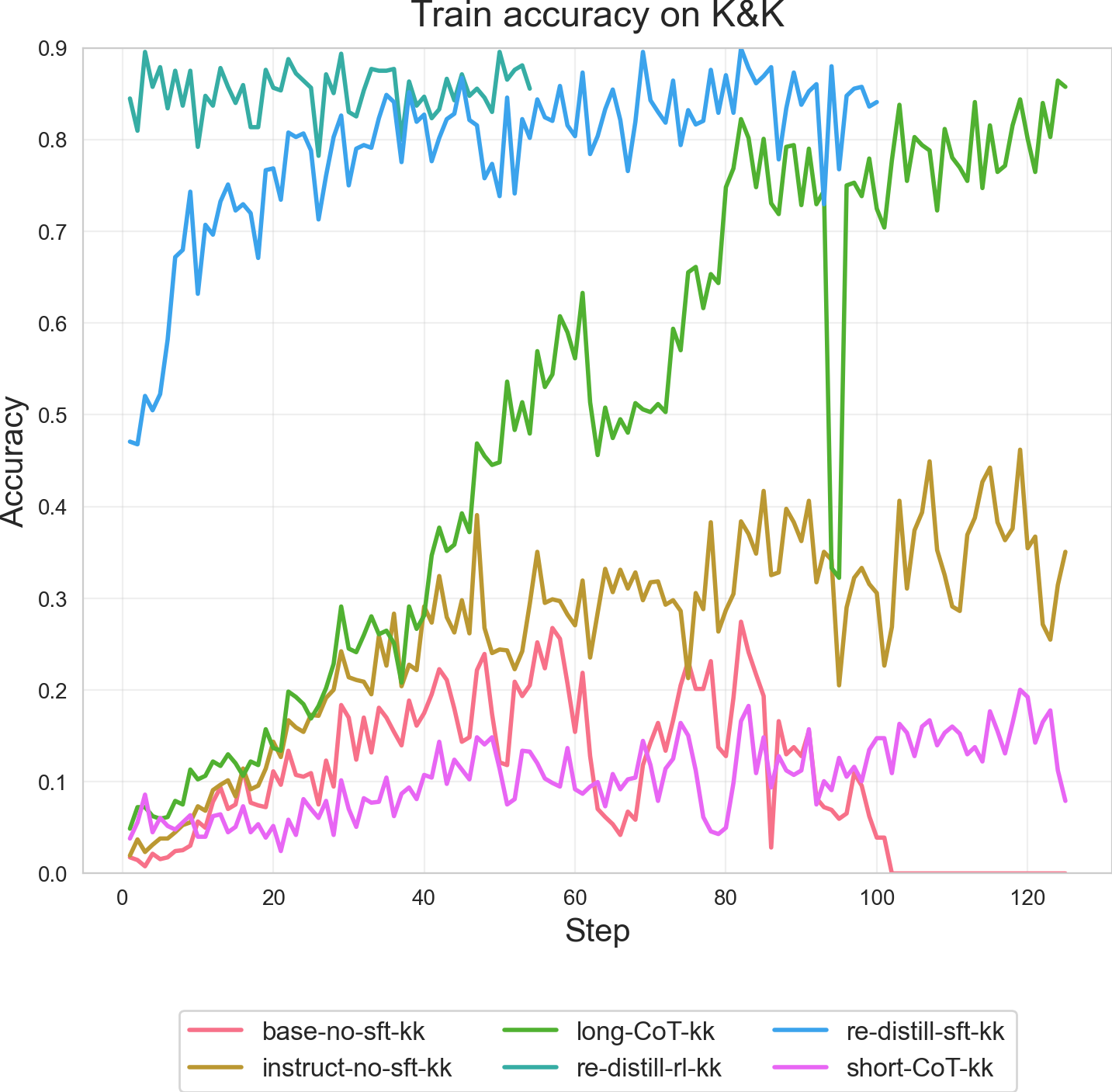}}
  \subfigure{\includegraphics[width=0.3\textwidth]{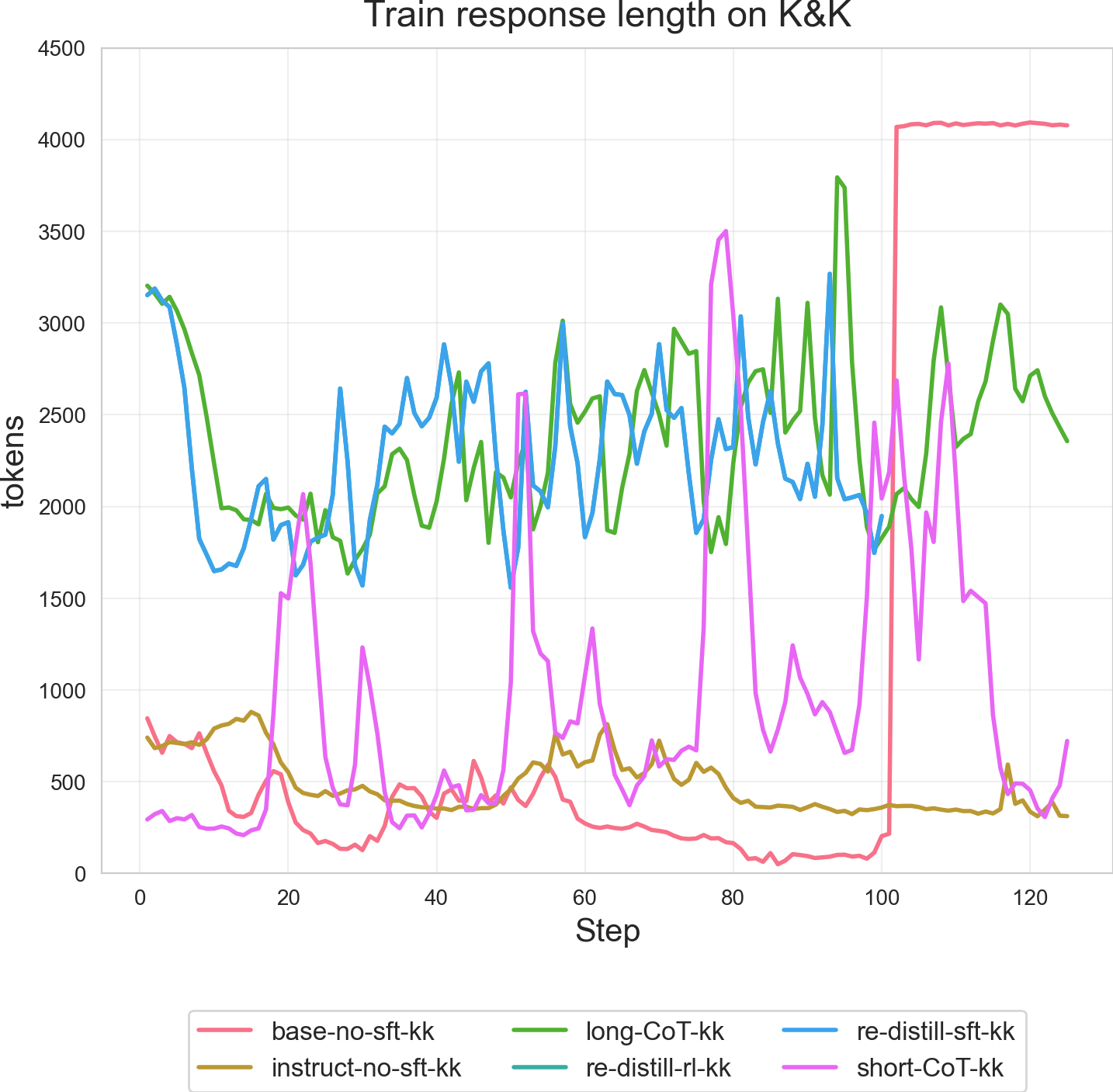}}
  \subfigure{\includegraphics[width=0.3\textwidth]{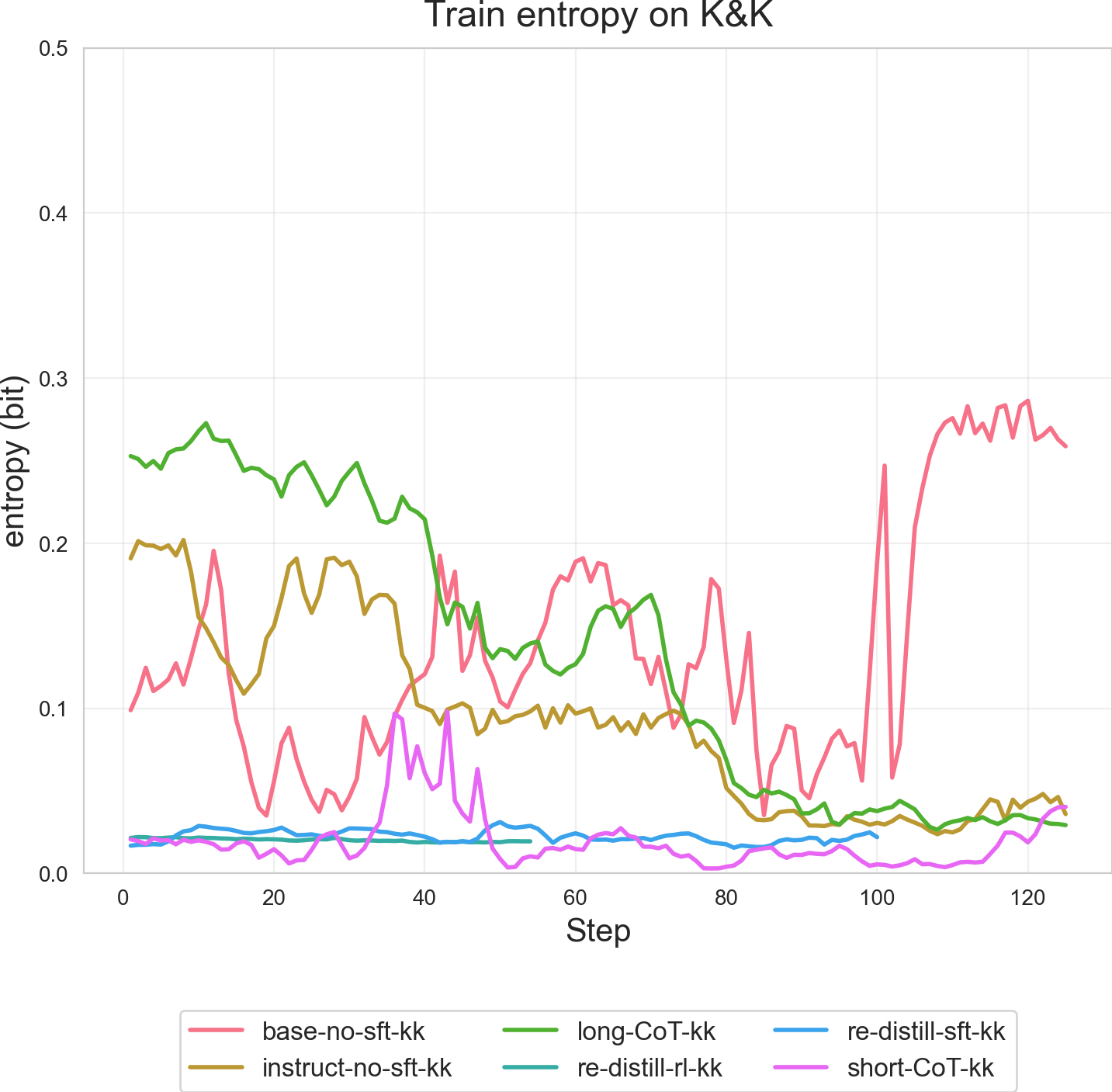}}
  \subfigure{\includegraphics[width=0.3\textwidth]{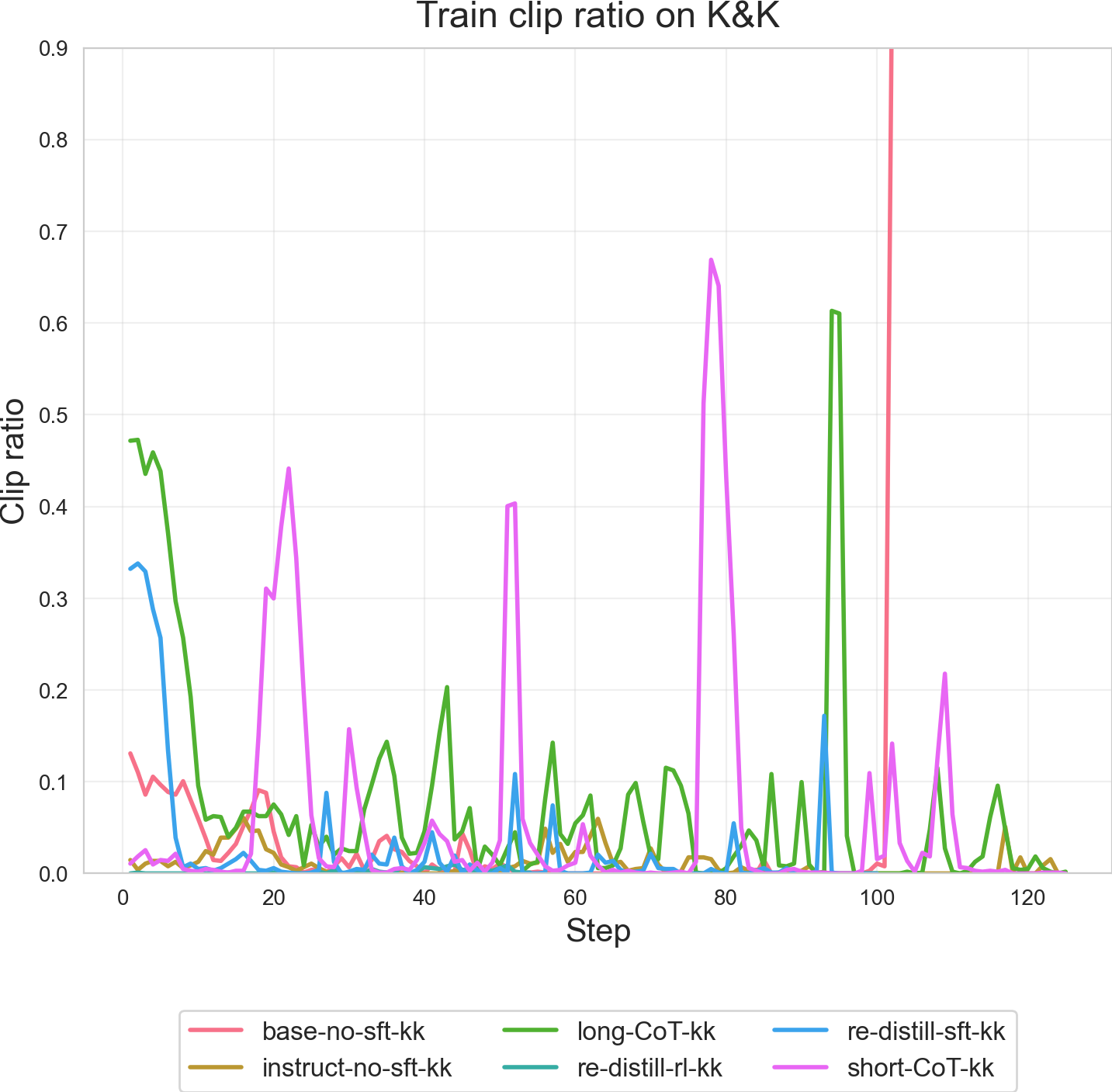}}
  \subfigure{\includegraphics[width=0.3\textwidth]{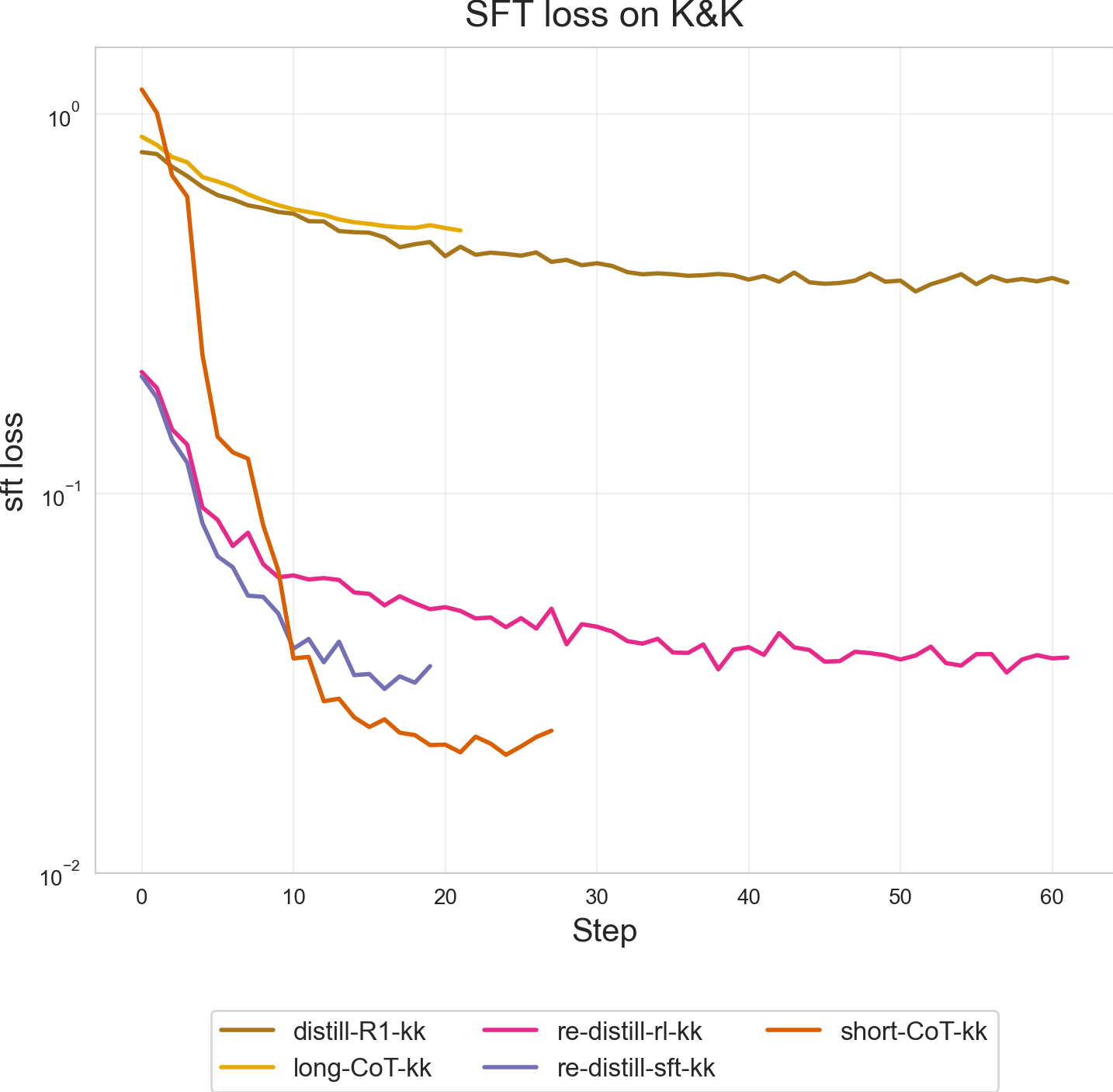}}

  \caption{Statistics of K\&K experiments. \textbf{Top Left}: Test accuracy in RL. \textbf{Top Middle}: Train accuracy in RL. \textbf{Top Right}: Training response length(token) in RL. \textbf{Bottom Left}: Training entropy in RL. \textbf{Bottom Middle}: Training clip ratio(exceeding max response length in RL). \textbf{Bottom Right}: SFT loss curve. }
  \label{fig:appendix-kk}
\end{figure}

We list detailed statistics for experiments on K\&K dataset in Figure \ref{fig:appendix-kk}.

\subsection{Detailed Statistics on MATH dataset} \label{subsec:appendix-stats-math}

\begin{figure}[h]
    \centering
    \subfigure{\includegraphics[width=0.3\textwidth]{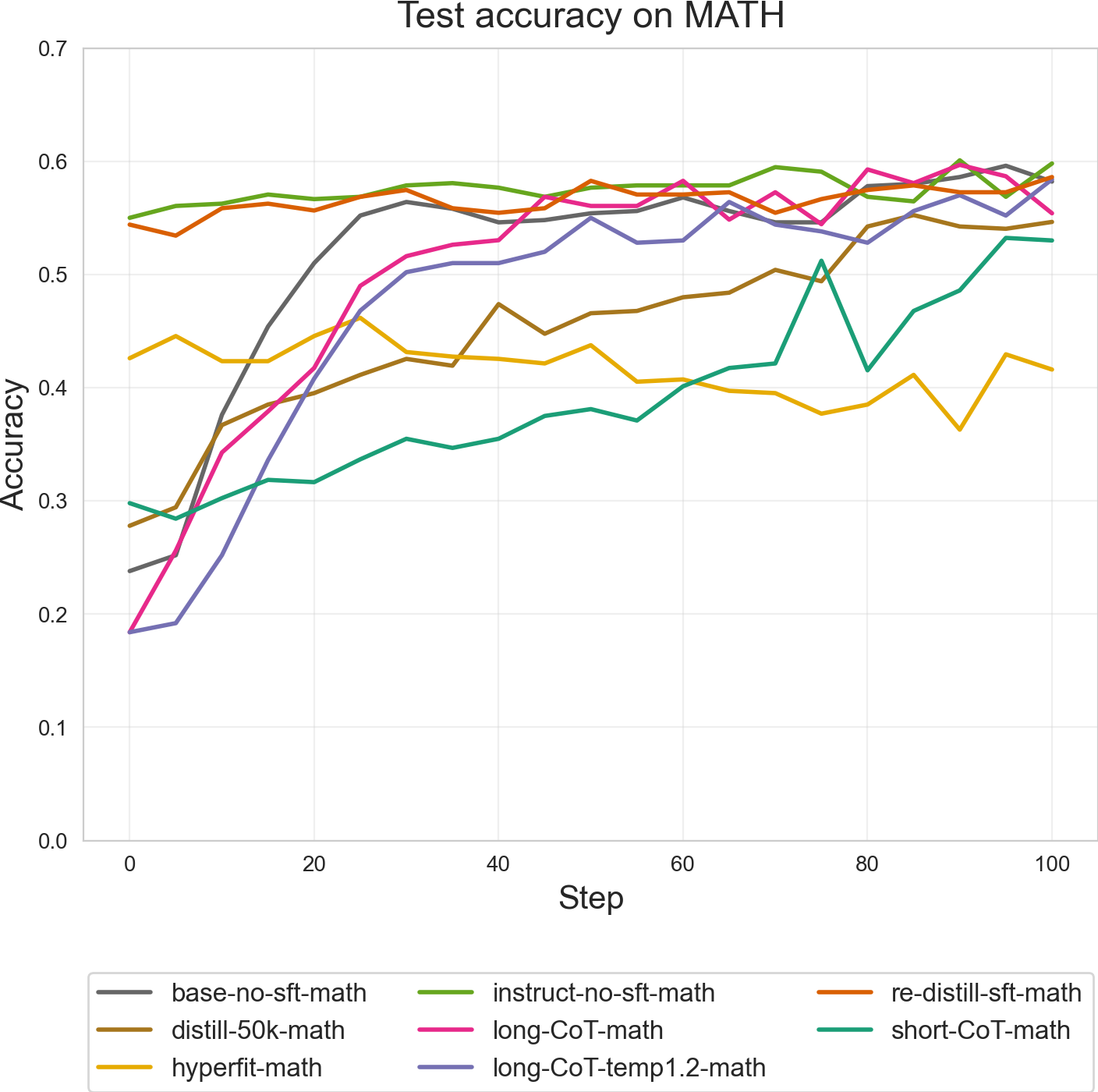}}
    \subfigure{\includegraphics[width=0.3\textwidth]{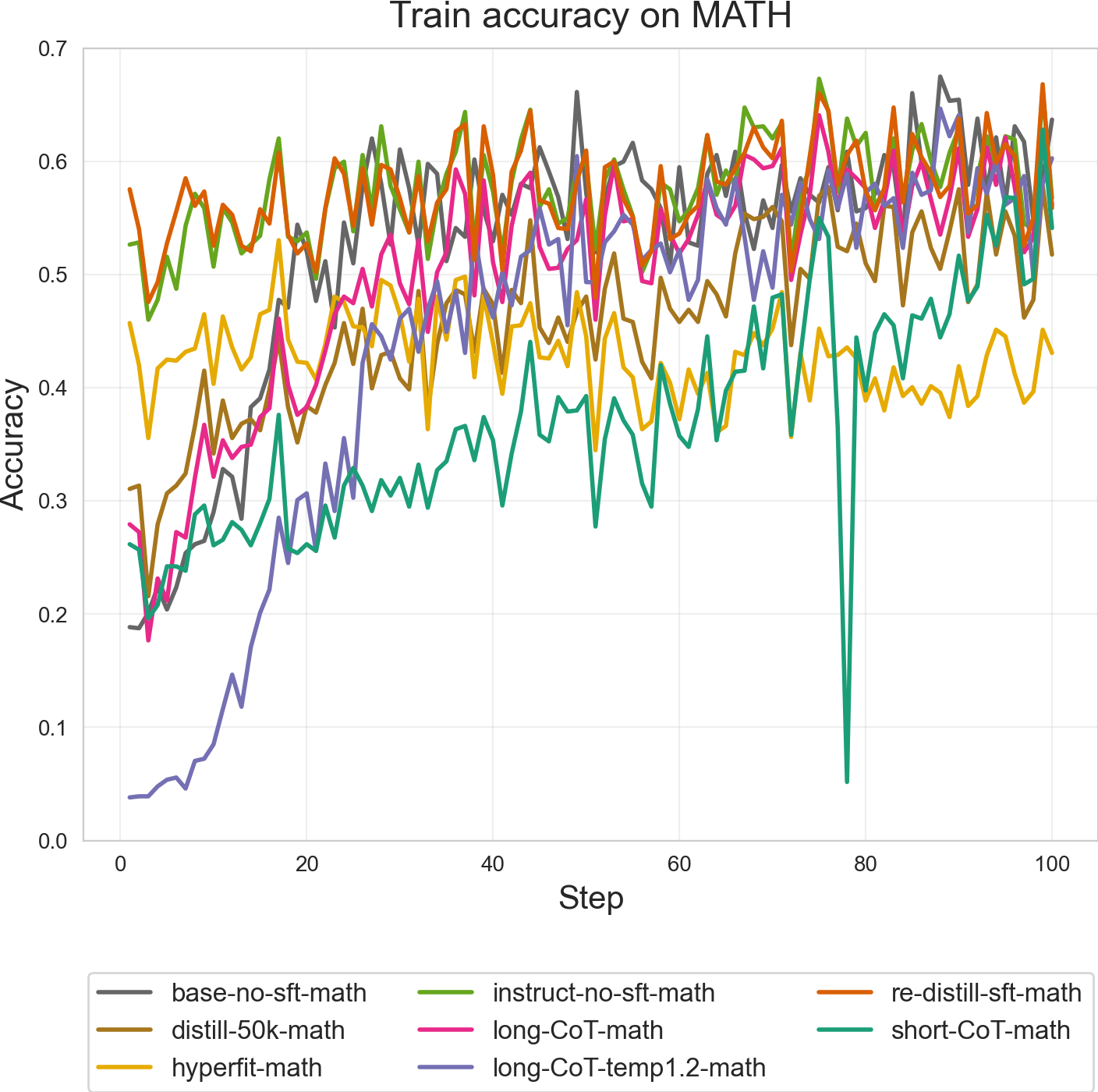}}
    \subfigure{\includegraphics[width=0.3\textwidth]{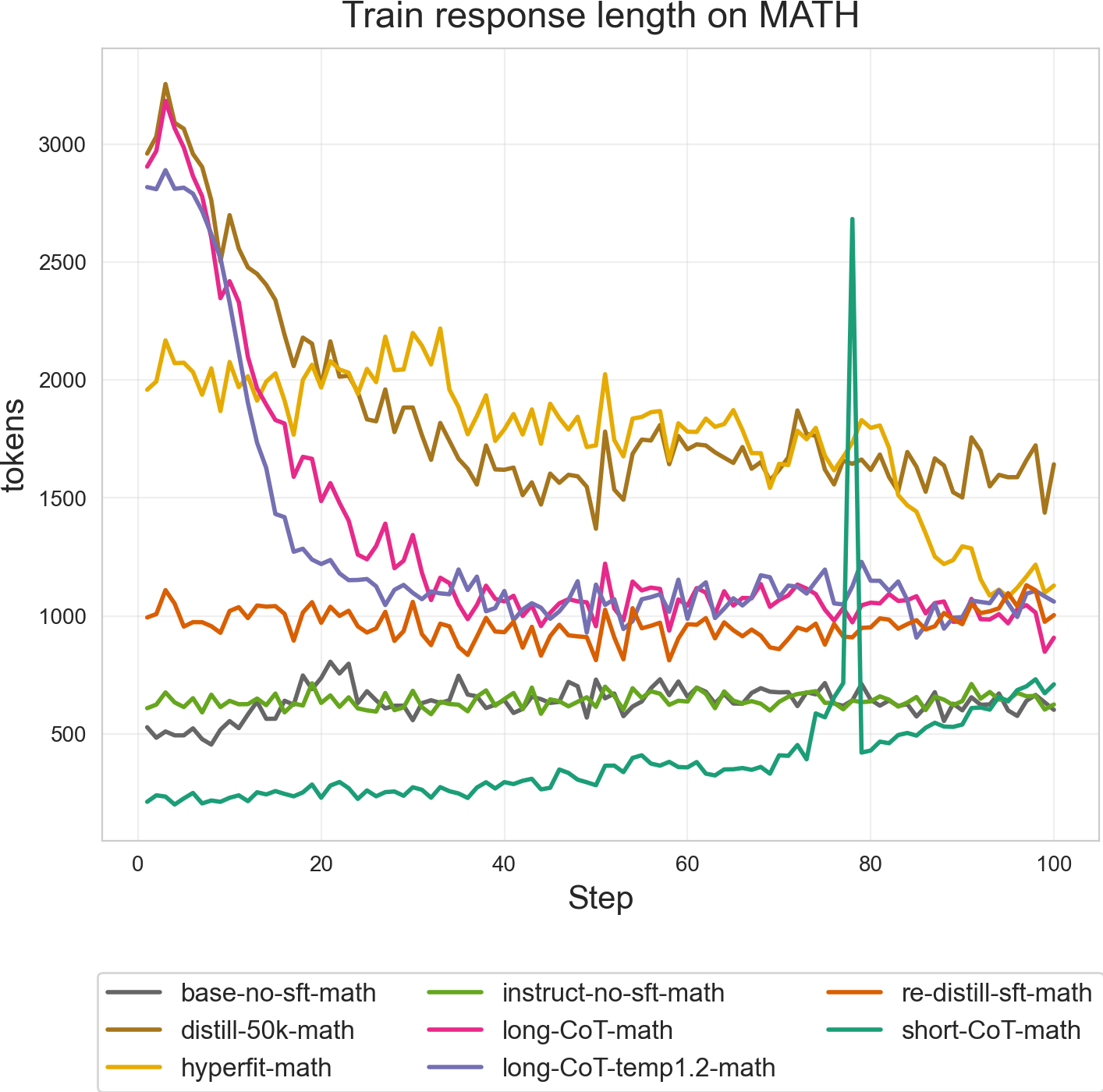}}
    \subfigure{\includegraphics[width=0.3\textwidth]{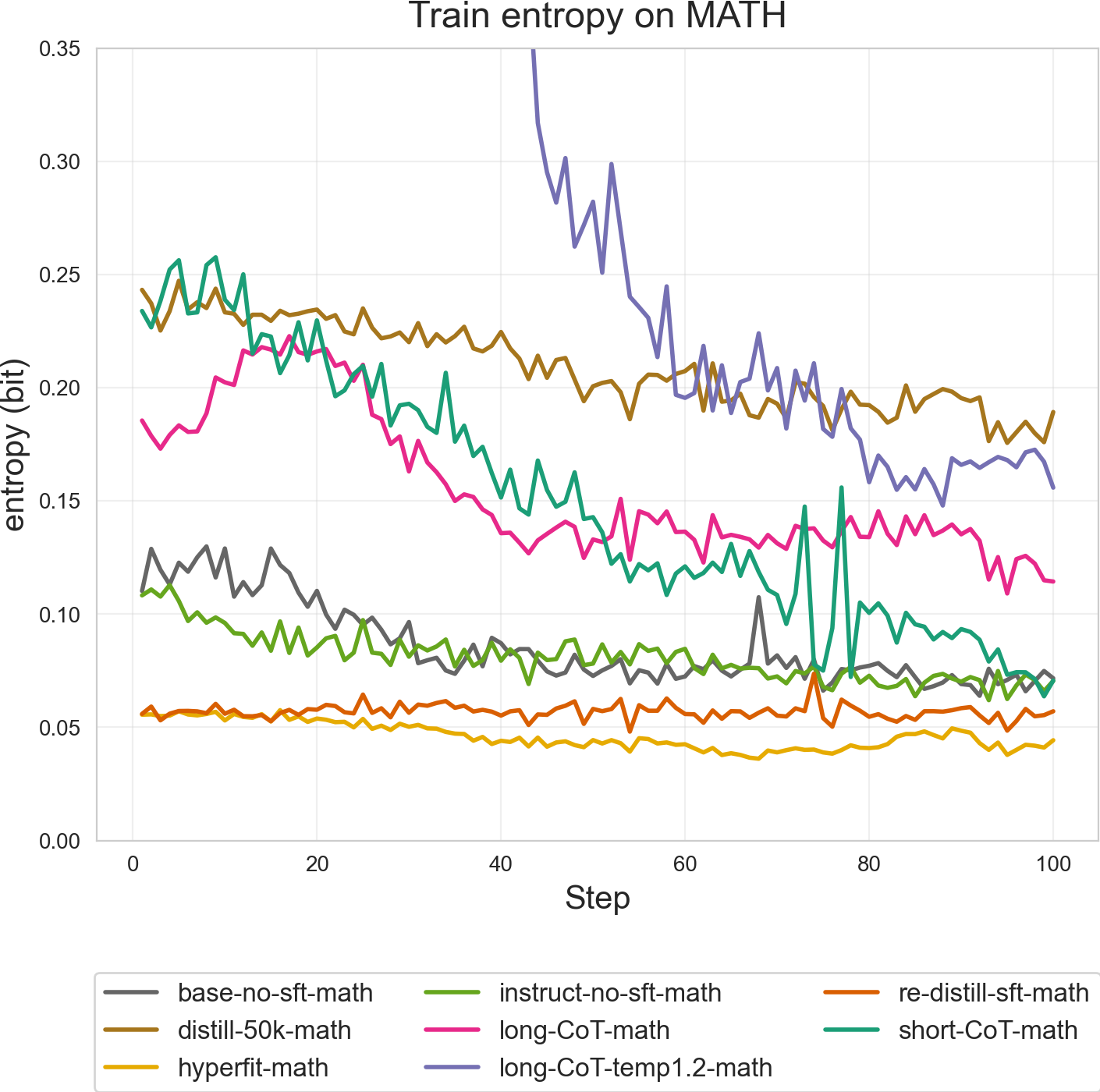}}
    \subfigure{\includegraphics[width=0.3\textwidth]{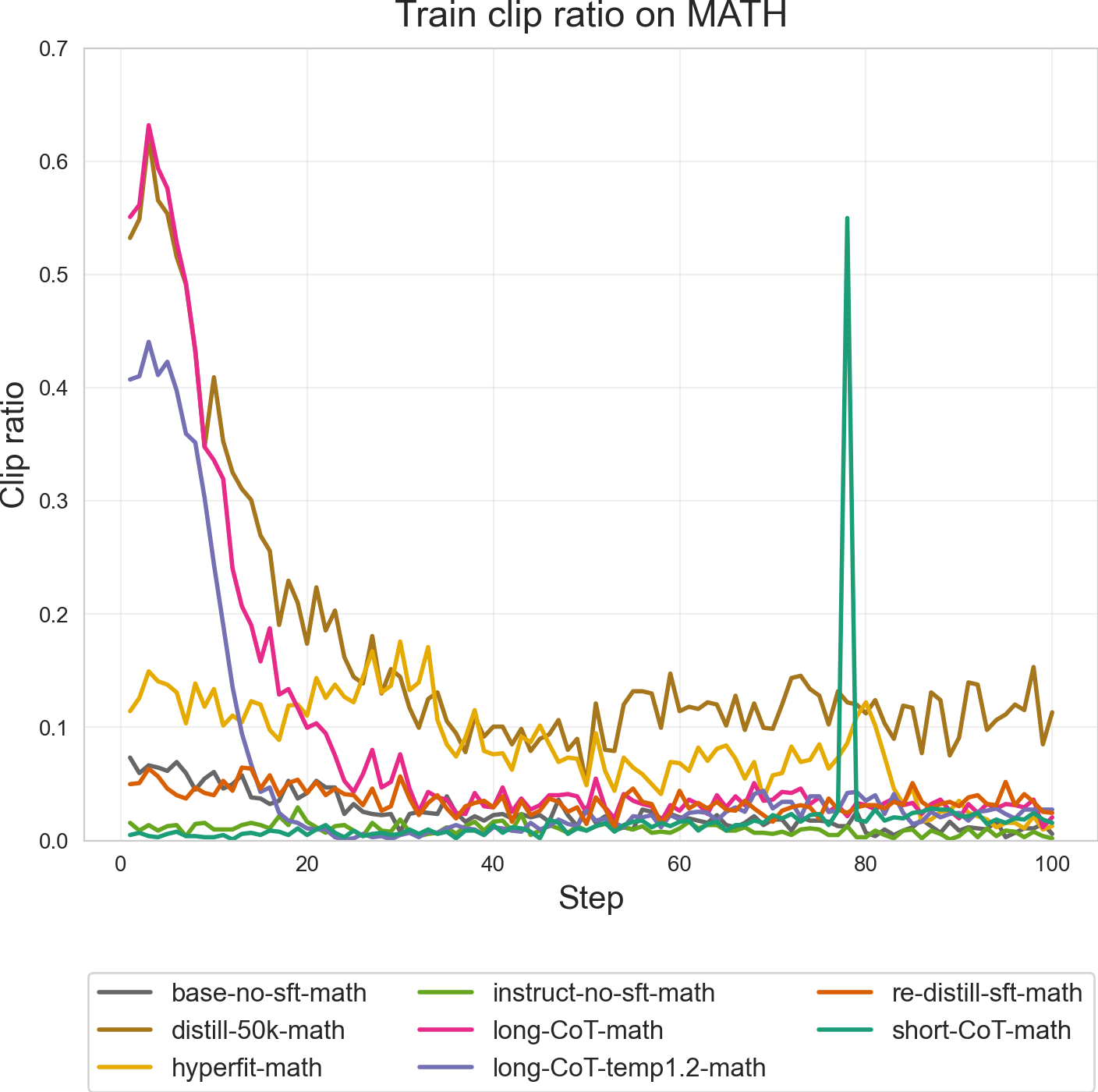}}
    \subfigure{\includegraphics[width=0.3\textwidth]{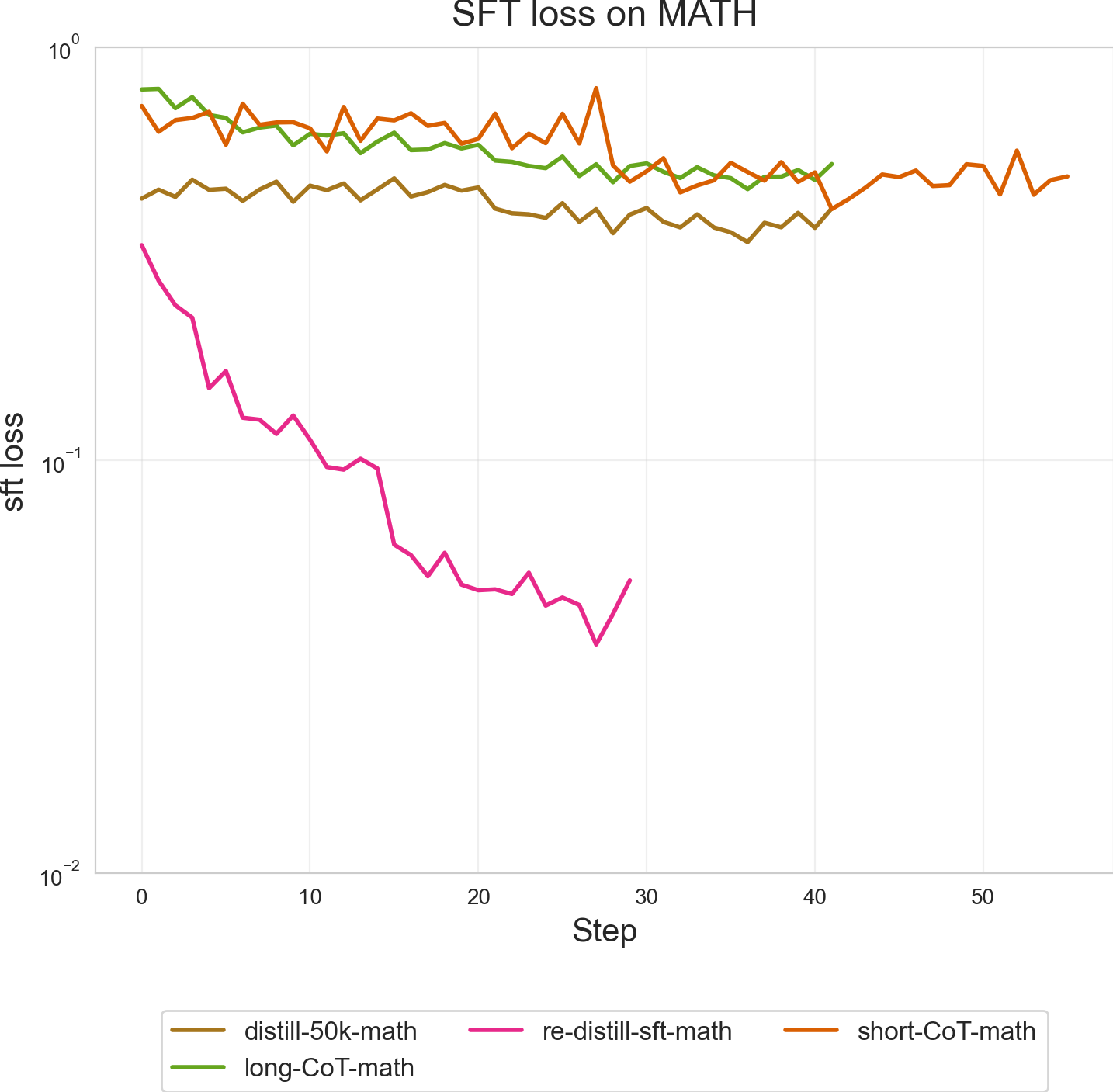}}
    \caption{Statistics of MATH experiments. \textbf{Top Left}: Test accuracy in RL. \textbf{Top Middle}: Train accuracy in RL. \textbf{Top Right}: Training response length(token) in RL. \textbf{Bottom Left}: Training entropy in RL. \textbf{Bottom Middle}: Training clip ratio(exceeding max response length in RL). \textbf{Bottom Right}: SFT loss in log scale.}
    \label{fig:appendix-math}
\end{figure}

We list detailed statistics for experiments on MATH dataset in Figure \ref{fig:appendix-math}. 

\subsection{Detailed Statistics on REASONING GYM dataset} \label{subsec:appendix-stats-rg}

\begin{figure}[h]
    \centering
    \includegraphics[width=0.6\textwidth]{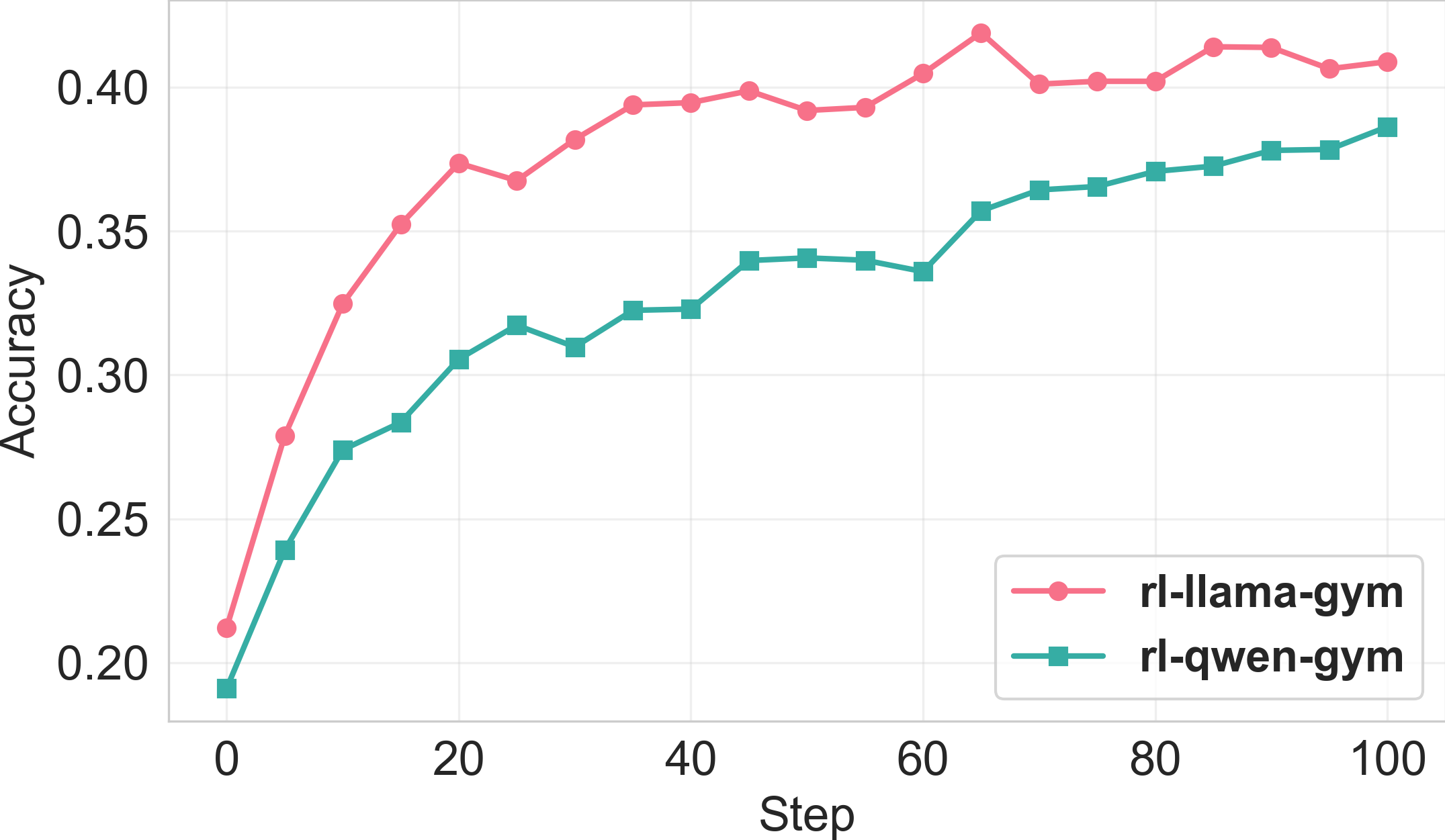}
    \caption{RL stage test accuracy for \texttt{rl-llama-gym} and \texttt{rl-qwen-gym}.}
    \label{fig:appendix-rg-rl-acc}
\end{figure}

\begin{figure}[h]
    \centering
    \subfigure{\includegraphics[width=0.95\textwidth]{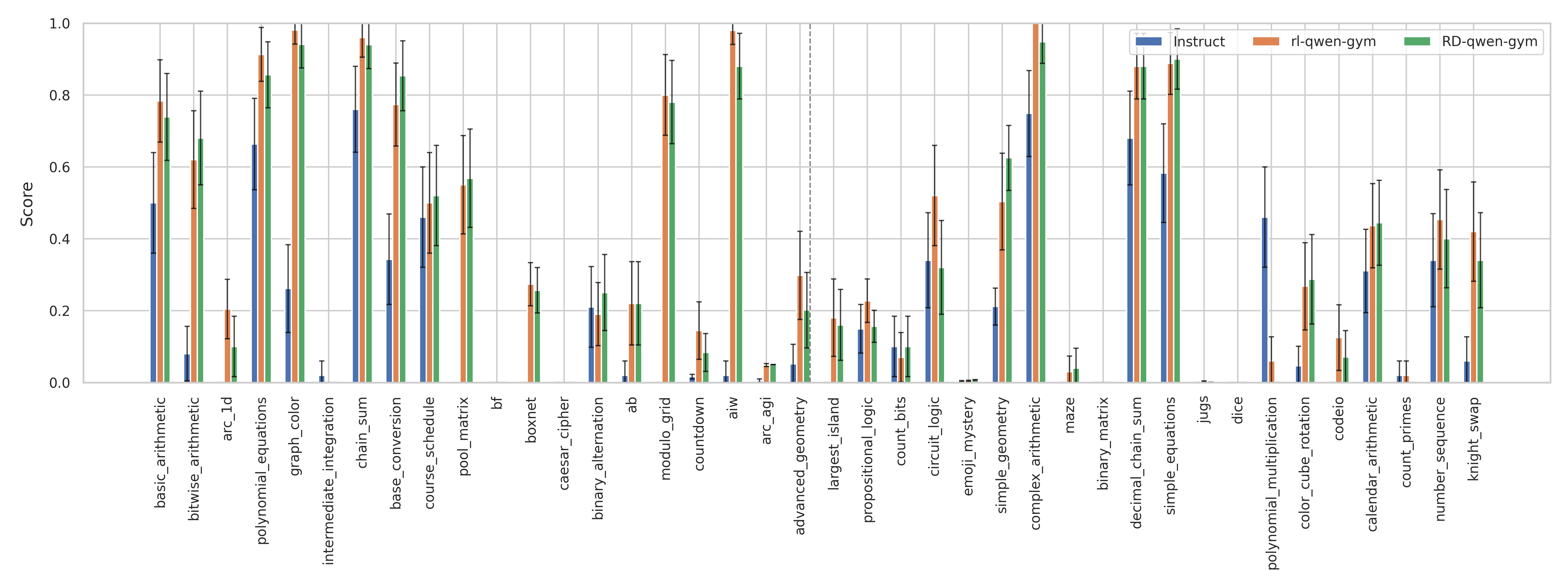}}
    \subfigure{\includegraphics[width=0.95\textwidth]{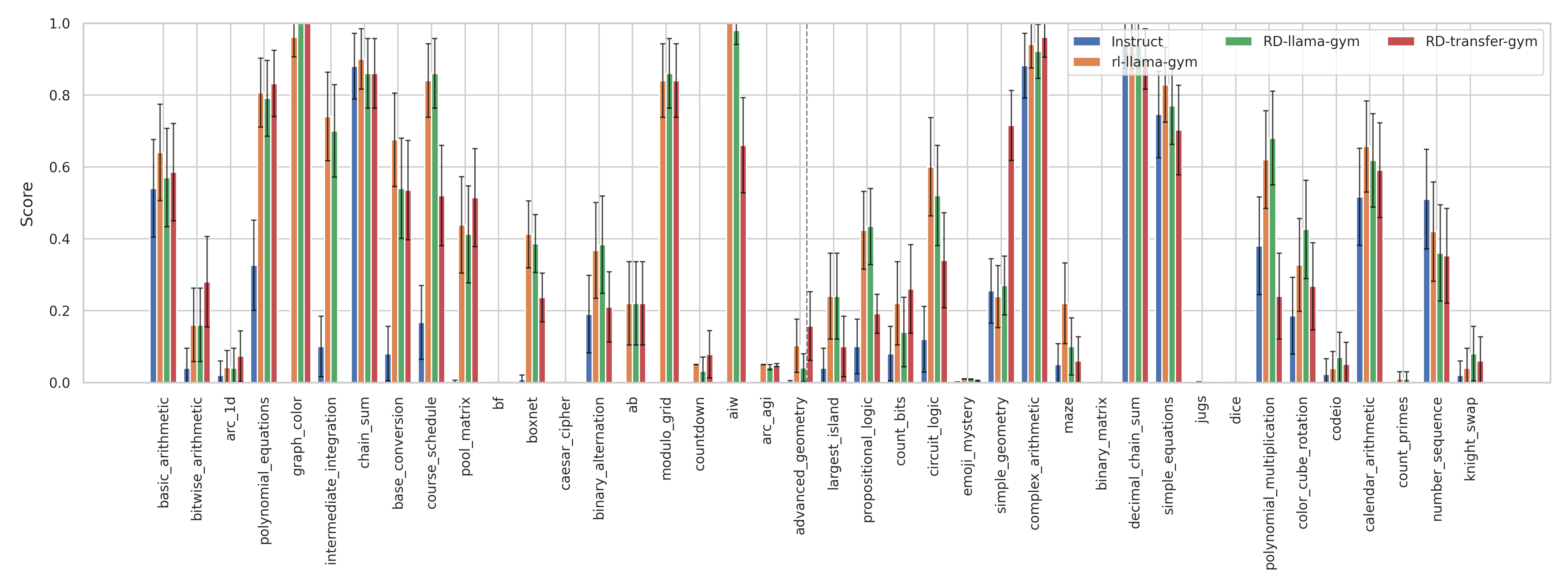}}
    \subfigure{\includegraphics[width=0.95\textwidth]{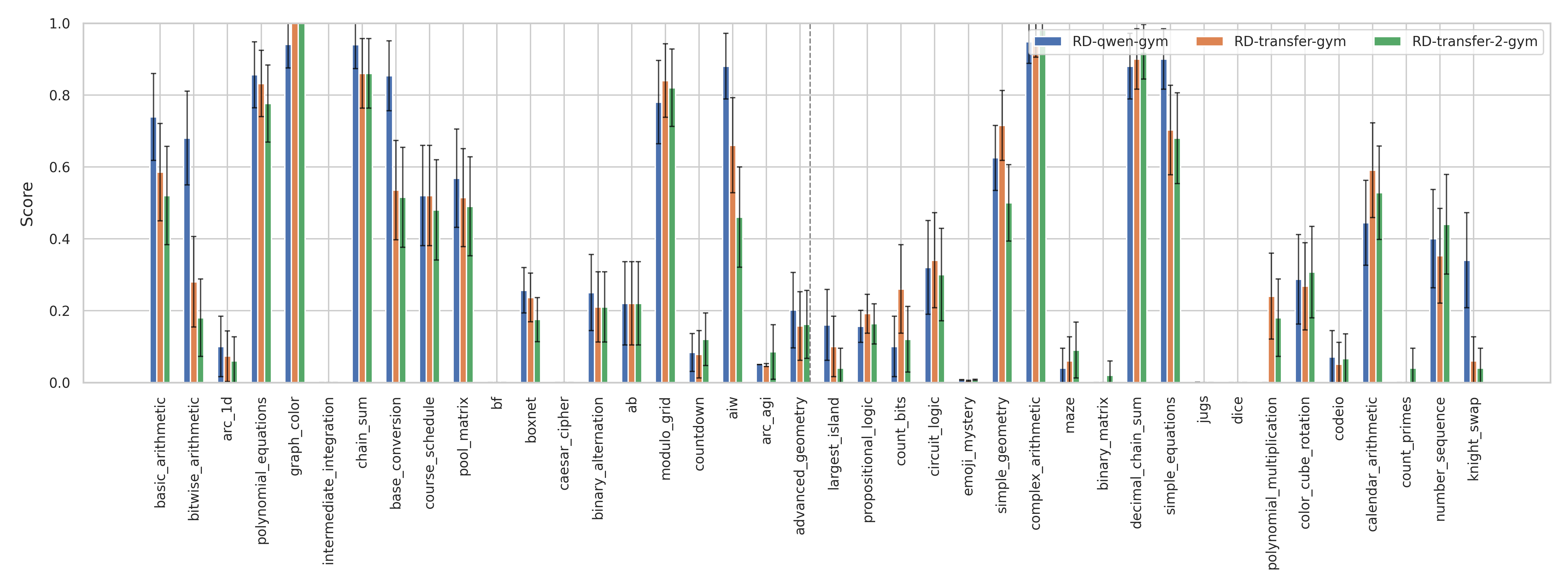}}
    \caption{Statistics of REASONING GYM experiments. Left side of black dash line is in-domain task; Right side is out-of-domain task. \textbf{Top}: Test accuracies of Qwen models. \textbf{Middle}: Test accuracies of Llama models. \textbf{Bottom}: Test accuracies of transferred models(\texttt{RD-transfer-gym} and \texttt{RD-transfer-2-gym})}
    \label{fig:appendix-rg}
\end{figure}

We show RL stage test accuracy for \texttt{rl-llama-gym} and \texttt{rl-qwen-gym} in Figure \ref{fig:appendix-rg-rl-acc}. We also list detailed accuracies for each in-domain and out-of-domain task in Figure \ref{fig:appendix-rg}. 

\end{document}